\documentclass[runningheads]{llncs}

\usepackage{eccv}

\usepackage{eccvabbrv}

\usepackage{graphicx}
\usepackage{booktabs}

\usepackage[accsupp]{axessibility}  

\usepackage[pagebackref,breaklinks,colorlinks,citecolor=eccvblue]{hyperref}

\usepackage{orcidlink}

\usepackage[dvipsnames]{xcolor}
\usepackage{multirow}
\usepackage{colortbl}
\usepackage{adjustbox}
\usepackage{bbm}

\usepackage{algorithm}
\usepackage{algorithmic}
\usepackage[accsupp]{axessibility}
\usepackage[normalem]{ulem}
\newcommand{\ours}{{FutureDepth}} 
\newcommand{\PNet}{{F-Net}} 
\begin{document}

\title{\ours : Learning to Predict the Future Improves Video Depth Estimation} 
\titlerunning{FutureDepth}



\author{Rajeev Yasarla, Manish Kumar Singh, Hong Cai,  Yunxiao Shi, Jisoo Jeong, Yinhao Zhu, Shizhong Han, Risheek Garrepalli, Fatih Porikli }
\authorrunning{Yasarla et al.}
\institute{Qualcomm AI Research\thanks{Qualcomm AI Research is an initiative of Qualcomm Technologies, Inc.} \\
\email{\tt\footnotesize	 \{ryasarla, masi, hongcai,  yunxshi, jisojeon, yinhaoz, shizhan, rgarrepa, fporikli\}@qti.qualcomm.com}\\}
\maketitle

\begin{abstract}
In this paper, we propose a novel video depth estimation approach, \ours, which enables the model to implicitly leverage multi-frame and motion cues to improve depth estimation by making it learn to predict the future at training. 
More specifically, we propose a future prediction network, F-Net, which takes the features of multiple consecutive frames and is trained to predict multi-frame features one time step ahead iteratively. In this way, F-Net learns the underlying motion and correspondence information, and we incorporate its features into the depth decoding process. Additionally, to enrich the learning of multi-frame correspondence cues, we further leverage a reconstruction network, R-Net, which is trained via adaptively masked auto-encoding of multi-frame feature volumes. 
At inference time, both F-Net and R-Net are used to produce queries to work with the depth decoder, as well as a final refinement network. 
Through extensive experiments on several benchmarks, i.e., NYUDv2, KITTI, DDAD, and  Sintel, which cover indoor, driving, and open-domain scenarios, we show that \ours~significantly improves upon baseline models, outperforms existing video depth estimation methods, and sets new state-of-the-art (SOTA) accuracy. Furthermore, \ours~is more efficient than existing SOTA video depth estimation models and has similar latencies when comparing to monocular models.
\vspace{-5pt}
\end{abstract}

\vspace{-15pt}
\section{Introduction}
\label{sec:intro} \vspace{-5pt}
Depth plays a critical role for 3D perception, in applications like autonomous driving, augmented reality/virtual reality (AR/VR), camera image and video processing, and robotics. While depth can be measured using LiDAR or Time-of-Flight (ToF) sensors, they are expensive, incur substantial power consumption, need to be calibrated, and can fail to produce correct measurements for certain surfaces (e.g., specular surfaces). As such, inferring depth based on camera images has become a cost effective and promising alternative. Traditional approaches~\cite{saxena2007depth, furukawa2010towards, newcombe2011dtam}, such as stereo vision and structure-from-motion, have been utilized to calculate depth, but have limited accuracy. Recently, by utilizing deep learning, researchers have achieved significantly more accurate depth estimation~\cite{eigen2014depth,fu2018deep,bhat2021adabins,Agarwal_2023_WACV,piccinelli2023idisc,yang2023gedepth}.

\begin{figure*}[t!]
    \vspace{-10pt}
    \centering
    \includegraphics[width=0.6\linewidth]{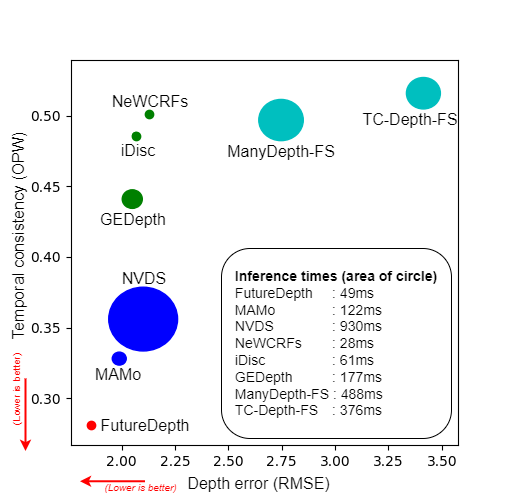} 
    \vskip -8pt 
    \caption{\small \ours~vs. existing SOTA in terms of depth accuracy (RMSE), temporal consistency (OPW~\cite{wang2023neural}), and runtime (on NVIDIA RTX-3080 GPU), on KITTI. We compare with monocular methods: NeWCRFs~\cite{yuan2022newcrfs}, iDisc~\cite{piccinelli2023idisc}, and GEDepth~\cite{yang2023gedepth}, cost-volume-based methods: ManyDepth~\cite{watson2021temporal} and TC-Depth~\cite{ruhkamp2021attention} (both fully supervised here), and video-based methods: MAMo~\cite{yasarla2023mamo} and NVDS~\cite{wang2023neural}. \ours~outperforms existing methods in terms of both accuracy and temporal consistency, and runs efficiently.}
    \label{fig:motivation}
    \vspace{-10pt}
\end{figure*}

Using a neural network to predict depth based on a single image, or monocular depth estimation, has been extensively studied due to simplicity of the setup and model efficiency~\cite{eigen2014depth,fu2018deep,bhat2021adabins,yuan2022newcrfs,Agarwal_2023_WACV,piccinelli2023idisc, yang2023gedepth}. 
This, however, does not take into account the consecutive video frames that are almost always available in many practical applications like autonomous driving and AR/VR.  
More recently, researchers have proposed different ways to exploit multiple frames for depth estimation. One common approach involves the employment of a cost volume, which assesses depth hypotheses and can be trained end-to-end within a neural network. Cost volumes can enable significant accuracy improvement, at the expense of significantly higher computational complexity and memory usage. Other researchers look into auto-regressive approaches, which does not require cost volumes but instead leverage alternative mechanisms such as recurrent neural network~\cite{zhang2019exploiting, patil2020don}, optical flow~\cite{eom2019temporally, xie2020video}, and/or attention~\cite{cao2021learning, wang2022less, yasarla2023mamo, wang2023neural}. 
While they can be more efficient than cost volume models, these methods still require costly operations to achieve SOTA accuracy. For instance, among latest works, MAMo~\cite{yasarla2023mamo} requires optical flow estimation, attention, and online gradient computation, while NVDS~\cite{wang2023neural} requires pairwise cross-attention between the target frame and every source frame to compute depth. Moreover, they do not consider predicting depths for consecutively frames jointly, and unable to learn underlying dynamic motion/trajectories of objects and corresponding spatial information, which
results in suboptimal temporal consistency.


In this paper, we propose a novel video depth estimation approach, \ours, which leverages future prediction and adaptive masked reconstruction to enable the model to learn and utilize key, multi-frame spatial and temporal features, while being computationally efficient.
More specifically, we take a representation learning approach and propose a multi-step Future Prediction Network, \PNet, which is trained to predict features of future frames that are one time step ahead based on the current given set of frame features, in an iteratively manner. This allows \PNet~to identify how pixel-wise scene and object features move across time, as it learns to predict the future. At inference time, \PNet~works together with the main encoder-decoder depth network and extracts useful motion features to enhance depth computation in the decoder.

In order to further enrich multi-frame correspondence understanding, we additionally propose a Reconstruction Network, R-Net, which is trained to perform Masked Auto-Encoding (MAE) on features of a consecutive set of frames with a learnable, adaptive masking strategy. This encourages R-Net to leverage critical scene features distributed across frames for reconstruction and thus, understand the multi-view correspondences. After training, R-Net parses the multi-frame features and generates key scene features to support the depth prediction. Note that this is different from existing video MAE methods (e.g.,~\cite{tong2022videomae, wang2023videomaev2}). Video MAE methods are used to pretrain the main encoder-decoder network, whereas we propose additional networks (\PNet~\& R-Net) work with the main network at inference (which can be any base depth network) to improve video depth estimation while maintaining computation efficiency.

Finally, we propose a small refinement network after the decoder, which further enhances the details of the predicted depth map. 


Our main contributions are summarized as follows:

\begin{itemize}
    \vspace{-10pt}
    \item We propose a novel approach, \ours, which leverages future prediction and adaptive masked reconstruction to enhance the model's ability to extract and exploit key, multi-frame motion and correspondence cues for video depth estimation.
    
    \item Our proposed Future Prediction Network, \PNet~adopts a multi-frame/time step future prediction loss based on auto-regressive sampling of future frames. This forces \PNet~to extract stronger motion, correspondence cues at inference time for better temporally consistent depth prediction. To the best of our knowledge, this is the first work to combine a multi-frame/time-step objective without teaching forcing and combine batch processing.
    
    \item We additionally propose a Reconstruction Network, R-Net, which is trained using learnable, adaptive masked auto-encoding of multi-frame features. Features generated by R-Net are also incorporated by the depth decoder to enrich scene understanding for better spatio-temporal aggregation. 
    Furthermore, we propose a small refinement network to improve the details of the final depth maps.

    \item We conduct extensive experiments on several depth estimation datasets: NYUDv2~\cite{silberman2012indoor}, KITTI~\cite{kitti}, DDAD~\cite{packnet}, and Sintel~\cite{butler2012naturalistic}. \ours~sets the new state-of-the-art (SOTA). 
    As shown in Fig.~\ref{fig:motivation}, \ours~has lower depth errors, is more temporally consistent, and runs faster than existing video depth methods and has similar latencies as compared to SOTA monocular models.
\end{itemize}

\begin{figure*}[t!]
    \centering
    \includegraphics[width=0.98\linewidth]{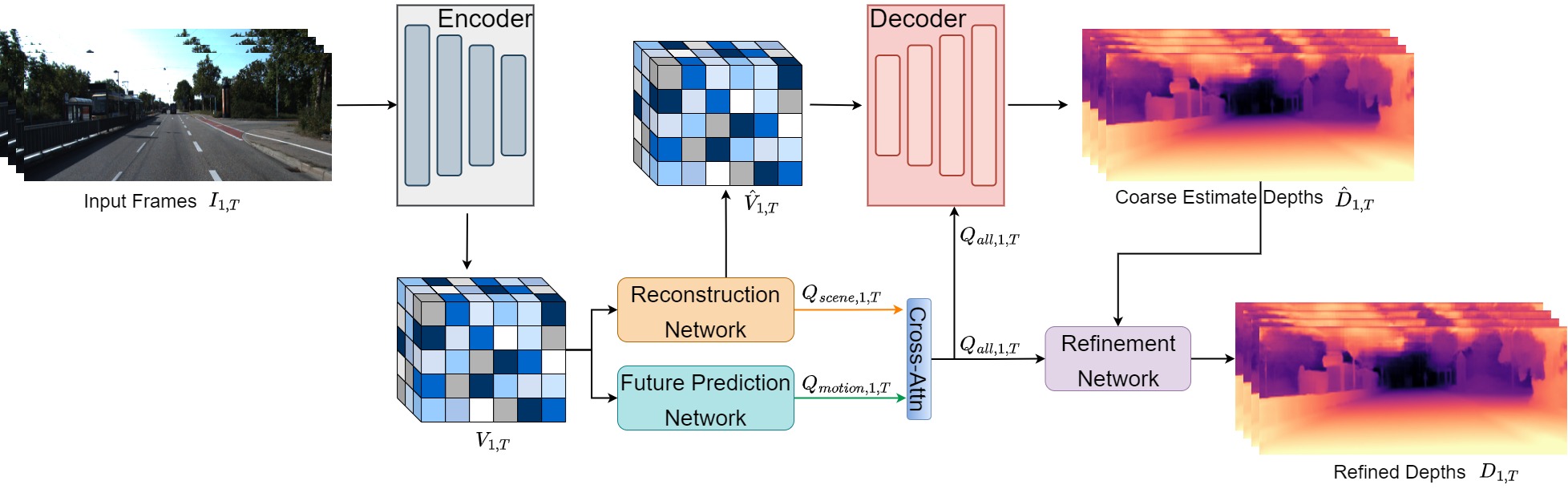} 
    \vskip -10pt 
    \caption{\small Our proposed \ours~method. Features of consecutive frames are extracted by the encoder and fed to the Future Prediction Network (F-Net) and Reconstruction Network (R-Net), which are trained using iterative future prediction and adaptive masked auto-encoding, respectively. At inference, features generated by F-Net and R-Net, $Q_{motion,1,T}$ and $Q_{scene,1,T}$, which contain key motion and correspondence cues, are integrated into the depth decoding process. Furthermore, these features are also utilized in a refinement process to improve the final depth map quality.}
    \label{fig:overview}
    \vspace{-11pt}
\end{figure*}

\vspace{-2pt}
\section{Proposed Approach: \ours}
\label{sec:method}
\vspace{-8pt}

Consider a batch of consecutive video frames, $I_{1,T} = \{I_1,I_2,...,I_{T}\}$, for which \ours~computes their depths jointly, $D_{1,T} = \{D_1,D_2,...,D_{T}\}$, where $T$ is the number of frames. The input frames are first fed into an image encoder individually. The image features are concatenated along the time and channel dimension to form a feature volume, $V_{1,T} = \{F_1,,F_2,...,F_{T}\} \in \mathbb{R}^{H'\times W'\times (T\cdot C)}$, where $F_t$ is feature map of $I_t$ extracted by the encoder, $H'$ and $W'$ are the height and width of the feature maps, and $C$ is the number of feature channels. 

In the proposed \ours~framework, we propose a Future Prediction Network, \PNet, which is trained with multi-step prediction of future features based on features of a given set of consecutive frames. In this way, \PNet~learns to capture the underlying motion information of the video frames and generates useful features to facilitate the depth prediction. To further enrich scene understanding, we additionally propose a Reconstruction Network, R-Net, which is trained via auto-encoding on the multi-frame features with a learnable masking scheme. In order to recover the original features, R-Net learns to seek available scene information distributed across frame and thus, implicitly identifies and utilizes multi-frame correspondences. After training, both networks work with the main encoder-decoder network, providing key, implicit motion and correspondence query features, $Q_{motion,1,T}$ and $Q_{scene,1,T}$, to enhance depth prediction. 
Fig.~\ref{fig:overview} provides an overview of our proposed \ours~approach. For conciseness, we drop the subscripts ``1'' and ``T'' on the query feature variables when the context is clear in the subsequent sections.

\vspace{-6pt}
\subsection{Future Prediction Network (\PNet)}
\vspace{-5pt}

\begin{figure*}[t!]
	\centering
	\includegraphics[width=0.9\linewidth]{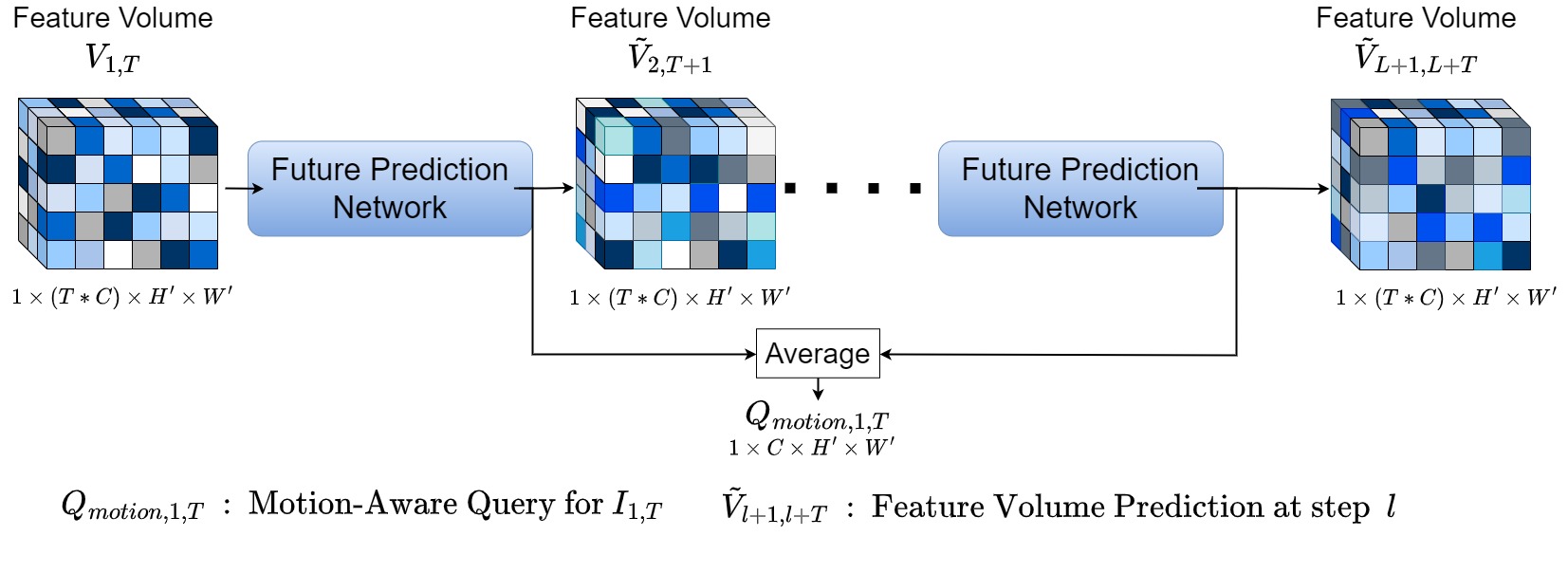} 
	\vskip -10pt 
	\caption{\small Future Prediction Network (\PNet).}
	\label{fig:future}
	\vspace{-8pt}
\end{figure*}

The goal of \PNet~is to capture useful, underlying motion cues to assist the video depth prediction process. 
To this end, we train \PNet~based on multi-step future prediction within the auto-regressive paradigm, i.e., we iteratively predict multi-frame features one step ahead with \textit{unrolling}. 
More specifically, \PNet~takes $V_{1,T}$ as input and predicts $\tilde{V}_{2,T+1}$. It then continues to take the currently predicted volume $\tilde{V}_{i,T+i-1}$ and predicts the one that is one step ahead $\tilde{V}_{i+1,T+i}$, until a prescribed number of steps, $L$, is reached. 
Our approach does not use teacher forcing, and the model is unrolled in an auto-regressive way. This means that when the network is predicting future values $\tilde{V}_{2,T+1}, \tilde{V}_{3,T+2}...\tilde{V}_{L+1,T+L}$, it does not take the actual future values $V_{2,T+1}, V_{3,T+2},...,V_{L,T+L-1}$ as input, when calculating the multi-step/multi-frame loss. This requirement compels the network to minimize error accumulation across time-steps during unrolling, thereby extracting more robust motion and correspondence cues. This process is illustrated in Fig.~\ref{fig:future}. 
Our work builds upon research in model-based reinforcement learning, where previous studies have explored the distinction between `observation-dependent' and `prediction-dependent' unrolling strategies \cite{hafner2019learning,hafner2023mastering,chiappa2017recurrent}. This also aligns with the concept of `covariate shift' explored in the context of imitation learning and teacher forcing \cite{alias2017z,spencer2021feedback}. To our knowledge, this is the first work to combine a multi-frame/time-step objective with batched processing, specifically aimed at improving the temporal consistency of dense perceptual tasks.

By using the feature volume of multiple frames to predict the entire volume one step ahead, \PNet~learns how the objects and scene move over time extracting stronger spatio-temporal information. Specifically, to predict the future feature at a certain pixel location, \PNet~needs to find the corresponding features that are available from the current and previous time steps. This essentially enables \PNet~to understand the underlying motion and multi-frame correspondences, as well as motion in longer contexts. 

\PNet~is trained with the following loss:\vspace{-8pt}
\setlength{\belowdisplayskip}{3pt} \setlength{\belowdisplayshortskip}{3pt}
\setlength{\abovedisplayskip}{6pt} \setlength{\abovedisplayshortskip}{3pt}
\begin{equation}\label{eqn:fut}
	\mathcal{L}_{F} = \sum_{i=1}^{L} || \tilde{V}_{1+i,T+i}-{V}_{1+i,T+i}||_2,
\end{equation}
where $\tilde{V}_{1+i, T+i-1}$ is the predicted volume for time steps $1+i$ to $T+i$.

\begin{figure}[t!]
    \vspace{-0pt}
	\centering
	\includegraphics[width=\linewidth]{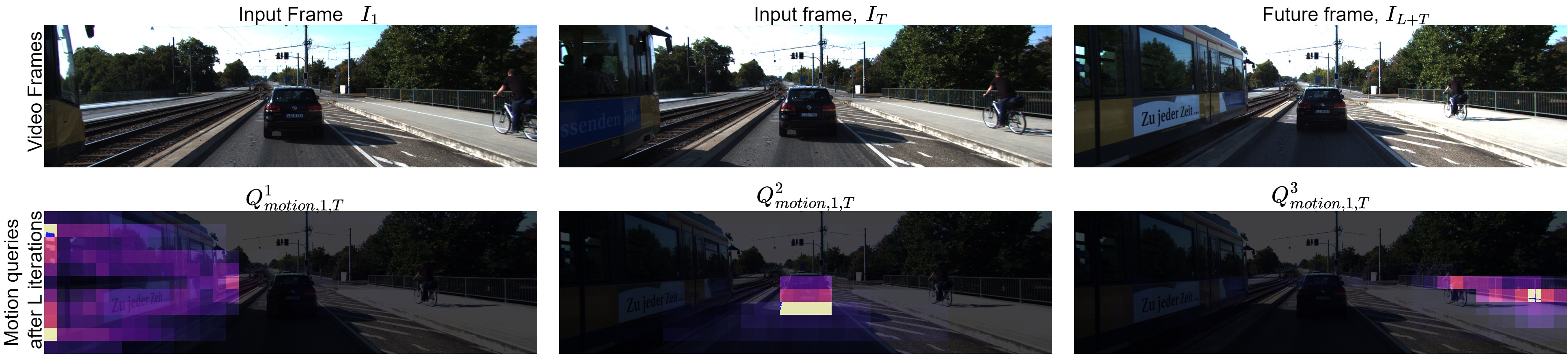} 
	\vskip -8pt 
	\caption{\small Example motion query $Q_{motion,1,T}$ generated using future prediction network. Here we show three example channels in $Q_{motion,1,T}$, with $T=4$, and $L=6$.}
	\label{fig:example-motion}
	\vspace{-1.0em}
\end{figure}

We use a transformer-based architecture for \PNet, which consists of convolutional and attention layers; more network details can be found in Sec.~\ref{sec:exp_setup}. To generate motion features that will be used for depth decoding, we use the last-layer features from all the prediction steps and average them to obtain $Q_{motion}$, which is fed into the depth decoder. 
Fig.~\ref{fig:example-motion} provides sample visualizations of $Q_{motion}$. We can see that moving objects are captured in $Q_{motion}$; particularly, extended motion understanding can be seen on the train and biker.


\vspace{-5pt}
\subsection{Reconstruction Network (R-Net)}
\vspace{-3pt}

We train R-Net using learnable, adaptive masked auto-encoding of the video feature volume. Suppose we have masks, $M_{1,T}=\{M_1,\dots,M_T\}$, which are derived from the input image features. We element-wise multiply the masks with the feature volume to generate the masked feature volume, $M_{1,T}\odot V_{1,T}$, which R-Net uses as as input and produces the reconstructed volume, $\hat{V}_{1,T}$. R-Net is trained using the following loss:
\begin{equation}\label{eqn:recon}
	\mathcal{L}_{R} = ||(1-M_{1,T})\odot (\hat{V}_{1,T}-{V}_{1,T})||_2+ \mathcal{L}_{D} ({D}_{1,T},D^{gt}_{1,T}), 
\end{equation}
which is the sum of the $L_2$ loss between the reconstructed and original feature volumes for masked locations, as well as a SILog loss~\cite{eigen2014depth}, $\mathcal{L}_{D}$, between predicted depths $D_{1,T}$ based on the reconstructed features and ground-truth depths $D^{gt}_{1,T}$.

We adopt a learned scheme to generate the masks based on the input image features.
Since no ground-truth masks are available, we train the mask generator using a SILog depth loss between predicted depths $D_{1,T}$ based on the masked features and ground-truth depths $D^{gt}_{1,T}$, with all the other network components frozen. We denote this loss as $\mathcal{L}_{A}$.

\begin{figure}[t!]
	\centering
	\includegraphics[width=\linewidth]{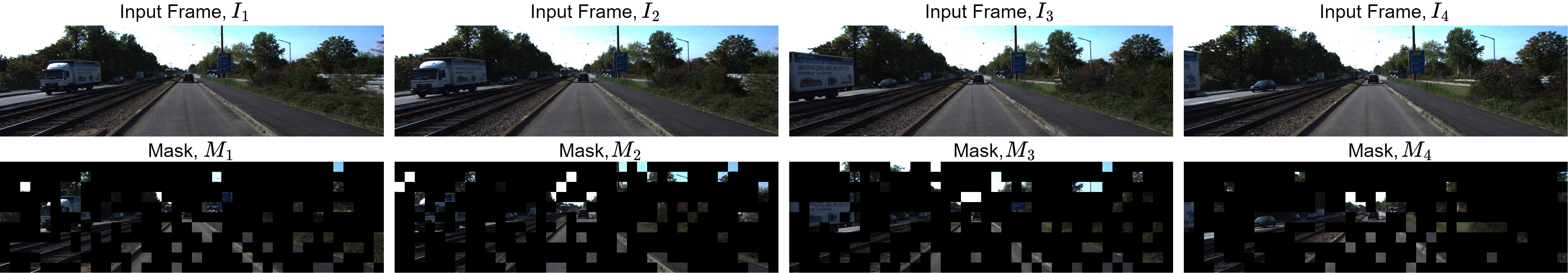} 
	\vskip -8pt 
	\caption{\small Generated masks over a sample input set of frames. Masks generated by adaptive sampler focuses on important objects like van, cars, tram, bus, railway tracks, and road boundaries etc. across the frames.}
	\label{fig:example-mask}
	\vspace{-10pt}
\end{figure}

We observe that the mask generator learns to mask the frames in a way that encourages R-Net to exploit multi-view correspondences. In particular, it masks out different parts of the same object across frames; for instance, see the white truck and the masks over it in Fig.~\ref{fig:example-mask}. As such, R-Net needs to utilize information distributed across frames in order to reconstruct the full feature volume.

\begin{figure}[t!]
	\centering
	\includegraphics[width=0.9\linewidth]{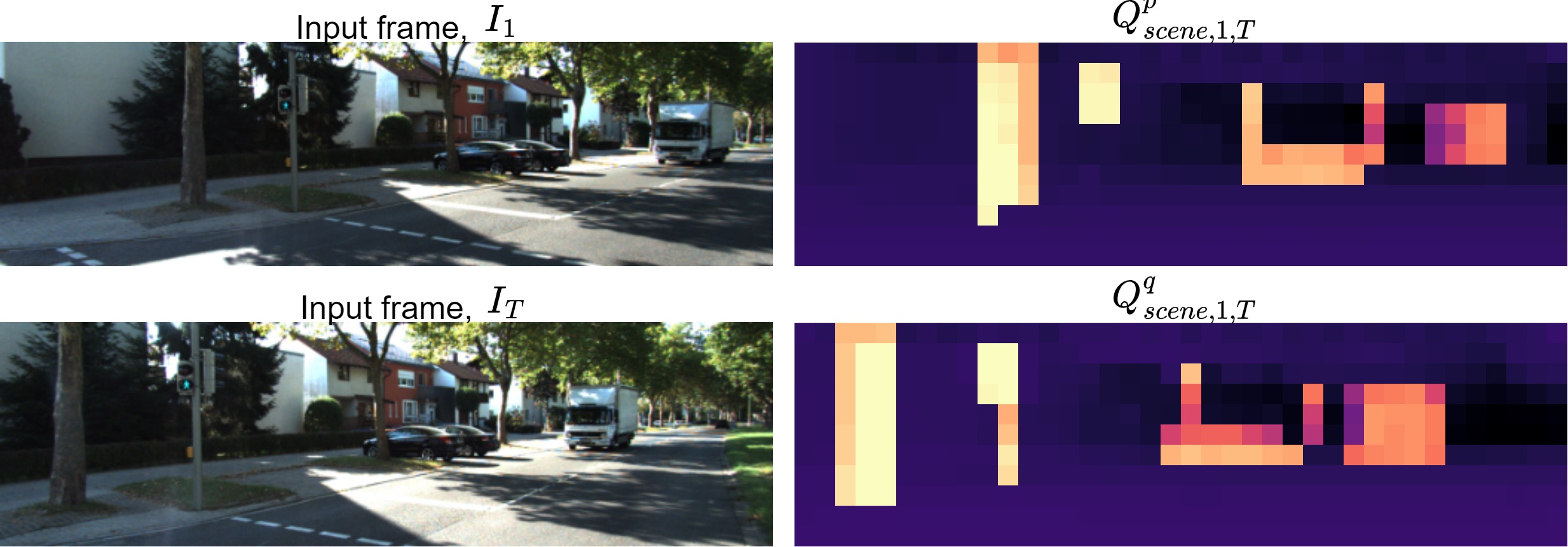} 
	\vskip -8pt 
	\caption{\small Sample $Q_{scene}$ generated by R-net. We show two sample channels $p$ and $q$ in $Q_{scene,1,T}$ for input frames $I_{1,T}\, (T=4)$.
 }
	\label{fig:example-qscene}
    \vspace{-15pt}
\end{figure}

We adopt the same architecture of \PNet~for R-Net. 
Once R-Net is trained, we use it to generate features, $Q_{scene}$, which contain important features of the scene and can be used to enhance depth estimation. More specifically, the input feature volume (not masked at inference) goes through convolutional and attention layers, and we use the output features from the last layer as $Q_{scene}$. More details of R-Net architecture can be found in Sec.~\ref{sec:exp_setup}. Fig.~\ref{fig:example-qscene} visualizes $Q_{scene}$. We see that in two sample channels, $Q_{scene}$ captures important foreground information (e.g., truck, parked cars, nearby tree) at different time steps.
 

\vspace{-7pt}
\subsection{Using the \PNet~and R-Net Features}
\vspace{-3pt}
The features provided by \PNet~and R-Net summarize key motion and scene information from the set of input frames. Given $Q_{motion}$ and $Q_{scene}$, we first perform cross-attention between them to produce $Q_{all}$. We expand the channel dimension of $Q_{all}$ by $T$, by repeating its $C$ channels. 

In the decoder, we employ transformer layers, where we combine $Q_{all}$ and the queries generated from the previous decoder layer's output, and process them at every decoder layer (more details are provided in the supplementary). In this way, we incorporate useful motion and multi-frame correspondence information into the depth prediction process.

In order to further improve the predicted depths from the decoder, we additionally design a progressive refinement network, consisting of self- and cross-attention layers. The refinement network takes depth maps as input and at the cross-attention layers, leverages $Q_{all}$ again to refine the depth map features. The refined version of the depth maps can be fed to the refinement network again to generate even further refined ones; we repeat this $N$ times in our pipeline. As we shall see, this helps improve the details of the depth maps.



\vspace{-7pt}
\subsection{Training}\label{sec:training}
\vspace{-3pt}
We first pretrain the \ours~encoder and decoder to perform depth prediction, without using \PNet, R-Net, and the refinement network. This training is supervised with a SILog loss between the predicted and ground-truth depths. 
After the encoder and decoder are trained, we train R-Net to perform feature volume reconstruction with random masking. In this step, we only use the $L_2$ loss to train R-Net, and do not update the encoder or decoder. 
The pretraining provides reasonable initial weights for several components in \ours, which is useful for a stable training of the entire system.

In the main training phase, we initialize both \PNet~and R-Net with the pre-trained R-Net weights, as they share the same architecture, as well as initialize the encoder and decoder with their pretrained weights. 
First, we freeze encoder and decoder, and train the adaptive mask generator, F-Net, and R-Net simultaneously.
We compute the $\mathcal{L}_{F}$, $\mathcal{L}_{A}$, and $\mathcal{L}_{R}$ losses and learn the weights for \PNet, mask generator, and R-Net, respectively. Finally, we freeze \PNet, mask generator, and R-Net, and train the encoder, decoder, and refinement network using the following loss:
\vspace{-5pt}
\begin{equation}\label{eqn:refine}
    \mathcal{L}_{D, final} = \frac{1}{NT}\sum_{i=0}^N\sum_{t=1}^T \mathcal{L}_{D}({D}^i_{t},D_{t}^{gt})
\end{equation}
where $L_D$ is the SILog loss~\cite{eigen2014depth} and the superscript $i$ indicates the refinement step. In this final step, $Q_{scene}$ and $Q_{motion}$ are cross-attended to generate $Q_{all}$, which is used in the decoding and refinement parts of the pipeline.

We refer readers to Algorithm~2 in the supplementary material for a more detail description of the training process. Note that after training, when performing inferences, we no longer need the mask generator and the inference pipeline is shown in Fig.~\ref{fig:overview}. Detailed description of the inference can be found in Algorithm~3 the supplementary. 


\section{Experiments}
\label{sec:experiments}
\vspace{-3pt}
We conduct extensive experiments to evaluate our proposed \ours~approach on large-scale public benchmarks and compare with existing state-of-the-art methods. We also
perform ablation studies to analyze different aspects of our proposed approach.

\subsection{Implementation and Experiment Setup}\label{sec:exp_setup}
\vspace{-3pt}
\noindent\textbf{Networks.} 
In \ours, we have a main an encoder-decoder depth network architecture, where the encoder can be any image backbones like ResNet~\cite{he2016deep}, ViT~\cite{dosovitskiy2020image}, and Swin~\cite{liu2021swin}, and the decoder consists of four Skip Attention Modules (SAM, modified from~\cite{Agarwal_2023_WACV}). 
We start from a single-frame baseline (designed by us) containing only an encoder and a decoder to compute depths, with 3 input channels and a single depth output channel. 
Then, we create a multi-frame baseline that processes $T$ frames as a batch, where their encoder outputs are concatenated to create a single feature volume with $T\times C$ channels. The feature volume is consumed by the decoder in its entirety, based on which the decoder predicts $T$ depth outputs in a batch manner. \ours~shares a similar base encoder-decoder architecture as the multi-frame baseline, but additionally includes our proposed F-Net, R-Net, and refinement network. In all our experiments, we use Swin-Large (Swin-L) as the encoder unless specified otherwise.

Both F-Net and R-Net consists of a self-attention layer, a convolutional layer that reduces the channel dimension from $T\cdot C$ to $C$, and four SAM layers. For R-Net, we use the queries from the last SAM layer as $Q_{scene}$. For F-Net, we use the last-layer queries from all prediction steps and average them to generate $Q_{motion}$. The mask generator consists of fully-connected layers and a softmax layer. Based on the softmax scores, we keep the top $r\times P$ patches and mask out the rest, where $r$ is the masking ratio and $P$ is total number of patches. The refinement network consists of one self-attention layer and two cross-attention layers. It takes depth maps as input and at the cross-attention layers, incorporates $Q_{all}$ to predict the improved depths.

\noindent\textbf{Hyperparameters.} We set number of video frames in a batch $I_{1,T}$ to be $T=4$. 
For each frame batch during training, we sample from ${1,2,3,4}$ to determine the interval between two consecutive frames. For instance, if the interval is 2, we sub-sample every other frame from the original sequence to form the batch. This allows the network to see more diverse motion ranges.
We set the number of iterations in future prediction to $L=T$ unless otherwise specified. We set the number of refinement step to $N=3$. 
In training, we sample the masking ratio from $r \in [0.6,0.9]$ to train R-Net. We set the initial learning rate to $4\times 10^{-5}$ and then linearly decrease it to $4\times 10^{-6}$.  In the pretraining stage, we train the encoder and decoder for 5 epochs, and subsequently, the R-Net for 3 epochs. In main training part, we train all the components of \ours~for 15 epochs. The total training takes about 2 days on 2 Nvidia A100 GPUs.

\noindent\textbf{Evaluation.} We use standard depth estimation metrics defined in~\cite{eigen2014depth}. In addition, we evaluate the temporal consistency of the predicted depths, using $aTC$ (lower is better) and $rTC$ (higher is better) from~\cite{li2021enforcing, yasarla2023mamo}, and OPW (lower is better) from~\cite{wang2023neural}. These metrics assess the prediction consistency across two frames by warping using optical flow.\footnote{Detailed mathematical definitions of the metrics can be found in the supplementary file.}

\subsection{Datasets}
\vspace{-3pt}

\noindent\textbf{NYUDv2~\cite{silberman2012indoor}.} This is a standard benchmark for indoor depth estimation tasks, containing 120K RGB-D videos captured from 464 indoor scenes. We follow the official Eigen training and test splits to evaluate our method, where 249 scenes are used for training and 654 images from 215 scenes are used for testing.

\noindent\textbf{KITTI~\cite{kitti}.} KITTI is one of the most commonly used benchmarks for outdoor depth estimation. We follow the Eigen training and test splits~\cite{eigen2014depth}, with 23,488 training images and 697 test images. We use the video 
(sub)sequences that correspond to the training and test image. These video frames are of size 375$\times$1241 and depth estimation is evaluated up to 80 meters.

\noindent\textbf{DDAD~\cite{packnet}.} Dense Depth for Autonomous Driving (DDAD) is a more recently introduced dataset featuring diverse urban driving scenarios with extended depth ranges. This dataset contains 12,650 samples for training and 3,950 samples for validation. We use the corresponding video (sub)sequences for training and running inferences. We follow the same setting introduced by~\cite{piccinelli2023idisc}, where images are cropped to 870$\times$1920 and the maximum depth is set to 150 meters.

\noindent \textbf{Sintel~\cite{butler2012naturalistic}.} MPI Sintel~\cite{butler2012naturalistic} consists of 23 synthetic sequences of open source animated films and captures open-domain scenarios. Following the protocol introduced by~\cite{ranftl2020towards,kopf2021robust}, we conduct zero-shot evaluation of our proposed \ours~and compare with state-of-the-art video depth estimation methods, which assesses model generalizability.


\begin{table}[t!]
\vspace{-2pt}
    \caption{\footnotesize Comparison with SOTA video-based models on NYUDv2 Eigen split and Sintel. OPW measures temporal consistency, as proposed in NVDS paper. FS means method is trained in fully-supervised fashion using ground-truth depth.  $\uparrow$ ($\downarrow$) means higher (lower) is better. 
    }
	\label{tab:NYUv2}
    \vspace{-10pt}
    \centering
    \resizebox{0.85\linewidth}{!}{
    \normalsize
    \begin{tabular}{l|ccc|ccc}
    \hline
    \multirow{2}{*}{Method}                                   & \multicolumn{3}{c|}{NYUDV2}      & \multicolumn{3}{c}{Sintel}                            \\ \cline{2-7} 
                            & \multicolumn{1}{c}{$\delta<1.25\uparrow$ } & \multicolumn{1}{c}{Abs Rel$\downarrow$} & OPW$\downarrow$ & \multicolumn{1}{c}{$\delta<1.25\uparrow$} & \multicolumn{1}{c}{Abs Rel$\downarrow$} & OPW$\downarrow$ \\ \hline
                 ST-CLSTM~\cite{zhang2019exploiting}        & {0.833}  & {0.131}  & 0.645  & {0.351}   & {0.517}   & 0.585  \\ 
                     FMNet~\cite{wang2022less}       & {0.832}     & {0.134}    &   0.387   & {0.357}     & {0.513}     &    0.521   \\ 
                      R-CVD~\cite{kopf2021robust}        & {0.886}     & {0.103}    &   0.394  & {0.521}     & {0.422}     &    0.475   \\ 
                      Many-Depth-FS~\cite{watson2021temporal}         & {0.865}     & {0.096}    &   0.428 & {0.492}     & {0.487}     &    0.540  \\ 
                      NVDS~\cite{wang2023neural}       & {0.950}     & {0.072}    &   0.364  & {0.591}     & {0.335}     &    0.424    \\ 
                      MAMo~\cite{yasarla2023mamo}      & {0.942}     & {0.074}    &   0.388  & {0.579}     & {0.358}     &    0.493     \\ \hline
                      Baseline (ours)       & {0.917}     & {0.093}    &   0.480   & {0.477}     & {0.504}     &    0.611   \\ 
                      \ours~(ours)         & {\textbf{0.981}}     & {\textbf{0.063}}    &   \textbf{0.303}  & {\textbf{0.623}}     & {\textbf{0.296}}     &    \textbf{0.392}  \\ \hline
    \end{tabular}
    }
    \vspace{-5pt}
\end{table}



\vspace{-3pt}
\subsection{Main Evaluation Results}
\vspace{-5pt}
\noindent\textbf{On NYUDv2.} In Table~\ref{tab:NYUv2}~(left), we evaluate our proposed \ours~approach and compare with existing SOTA on NYUDv2. It can be seen that \ours~outperforms latest existing SOTA video depth estimation models (e.g., MAMo, NVDS) and sets the new SOTA accuracy. In particular, we reduce the depth error (in terms of Abs Rel) by more than 12\% when comparing to NVDS and MAMo. It is also noteworthy that we improve temporal consistency (measured by OPW) by more than 16\%.

\begin{table*}[t!]
    \small
    \caption{Quantitative results on KITTI (Eigen split). $\dagger$ indicates methods using multiple networks to estimate depth. The methods are ordered in each group based on Abs Rel. MF means multi-frame methods. SF means single-frame methods.  }
	\label{tab:KITTI}
    \centering
    \vspace{-6pt}
    \adjustbox{max width=1\textwidth}
    {
    \begin{tabular}{ll|l|ccccc}
    \hline
        Type & Method & Encoder & \quad Abs Rel$\downarrow$ \quad  & \quad Sq Rel$\downarrow$ \quad & \quad RMSE$\downarrow$ \quad & \quad $\text{RMSE}_{log}\downarrow$ \quad & \quad $\delta<1.25\uparrow$ \quad  \\ \hline
        \multirow{6}{*}{SF} & AdaBins~\cite{bhat2021adabins} &  EfficientNet &0.058	& 0.190	& 2.360	&0.088	&0.964		\\	
        & BinsFormer~\cite{li2022binsformer} & Swin-L &0.052	&0.151	&2.098	&0.079&0.975			\\
        &NeWCRFs~\cite{yuan2022newcrfs} & Swin-L & 0.052 & 0.155 & 2.129 & 0.079 & 0.974			\\	
        & PixelFormer~\cite{Agarwal_2023_WACV} & Swin-L & 0.051 & 0.149 & 2.081 & 0.077 & 0.976	 \\ 
        & iDisc~\cite{piccinelli2023idisc} & Swin-L & 0.050 & 0.145 & 2.067 & 0.077 & 0.977	 \\ 
        & GEDepth~\cite{yang2023gedepth} & \cite{li2022depthformer} & 0.048 & 0.142 & 2.050 & 0.076 & 0.976	 \\ 
        \hline
        \multirow{14}{*}{MF} & FlowGRU~\cite{eom2019temporally} & \cite{eom2019temporally} & 0.112 & 0.700 & 4.260 & 0.184  & 0.881 \\ 
        & RDE-MV~\cite{patil2020don} &  ResNet18$\dagger$ & 0.111 & 0.821 & 4.650 & 0.187  & 0.821 \\
        & STAD~\cite{lee2021stad} & \cite{liu2019neural}$\dagger$ & 0.109 & 0.594 & 3.312 & 0.153  & 0.889  \\
        & Patil \textit{et.al.}~\cite{patil2020don} &  ConvLSTM$\dagger$ & 0.102 & -- & 4.148 & --  & 0.884  \\
        & ST-CLSTM~\cite{zhang2019exploiting} & ResNet18 & 0.101 & -- & 4.137 & --  & 0.890  \\ 
        & NeuralRGB~\cite{liu2019neural} &  CNN-based$\dagger$ & 0.100 & -- & 2.829 & --  & 0.931  \\ 
        & Cao \textit{et.al.}~\cite{cao2021learning} &  -- & 0.099 & -- & 3.832 & --  & 0.886 \\ 
        & FMNet~\cite{wang2022less} &  ResNeXt-101 & 0.099 & -- & 3.744 & 0.129  & 0.888  \\  
        & Flow2Depth~\cite{xie2020video} &  \cite{mayer2016large}$\dagger$ & 0.081 & 0.488 & 3.651 & 0.146  & 0.912  \\ 
        & TC-Depth-FS~\cite{ruhkamp2021attention} &  ResNet50 & 0.071 & 0.330 & 3.222 & 0.108  & 0.922  \\
        & ManyDepth-FS~\cite{watson2021temporal} &  ResNet50 & 0.069 & 0.342 & 3.414 & 0.111  & 0.930  \\
        & ManyDepth-FS~\cite{watson2021temporal} &  Swin-L & 0.060 & 0.248 & 2.747 & 0.099  & 0.955  \\
         & NVDS~\cite{wang2023neural} &  DPT-L\cite{ranftl2021vision} & 0.052 & 0.159 & 2.101 & 0.077 & 0.976  \\
        & MAMo~\cite{yasarla2023mamo} &  Swin-L & 0.049 & 0.130 & 1.989 & 0.072 & 0.977\\  \hline
        \multirow{8}{*}{\begin{tabular}[l]{@{}l@{}} MF\\ (Ours) \end{tabular}} &  \multirow{4}{*}{ Baseline} &   ResNet34 & 0.063 & 0.219 & 2.521 & 0.098 & 0.957    \\
        &   &  Swin-B & 0.055 & 0.162 & 2.163 & 0.082 & 0.973   \\
        &   &  Swin-L & 0.053 & 0.154 & 2.094 & 0.079 & 0.975  \\
        &   & Dinov2 (ViT-L) & 0.051 & 0.141 & 2.064 & 0.076 & 0.979   \\ \cline{2-8}
        &  \multirow{4}{*}{ \ours~} &  ResNet34 & 0.054 & 0.179 & 2.016 & 0.087 & 0.965  \\
        &   &  Swin-B & 0.049 & 0.129 & 1.998 & 0.077 & 0.976   \\
        &   &  Swin-L & 0.044 & 0.119 & 1.920 & 0.068 & 0.983     \\ 
        &   &   Dinov2 (ViT-L) & \textbf{0.041} & \textbf{0.117} & \textbf{1.856} & \textbf{0.066} & \textbf{0.984}  \\
        \hline   
    \end{tabular}
    }
    \vspace{-15pt}
\end{table*}


\begin{figure*}[t!]
\begin{center}$
\centering
\begin{tabular}{c c c c c c}
\scriptsize{Input} & \scriptsize{iDisc} & \tiny{ManyDepth-FS} & \scriptsize{NVDS} & \scriptsize{MAMo} & \scriptsize{FutureDepth} \\
\vspace{-0.05cm}
\hspace{-0.3cm} \includegraphics[width=1.98cm]{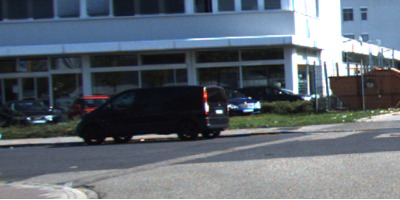} 
& \hspace{-0.15cm} \includegraphics[width=1.98cm]{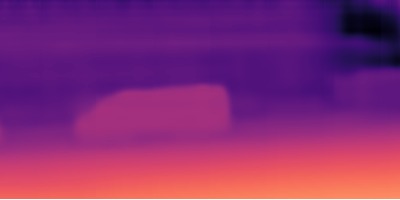}
& \hspace{-0.15cm} \includegraphics[width=1.98cm]{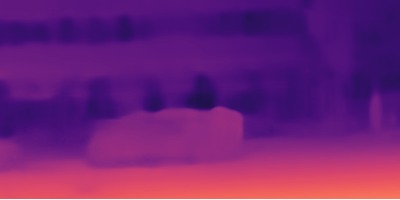}
& \hspace{-0.15cm} \includegraphics[width=1.98cm]{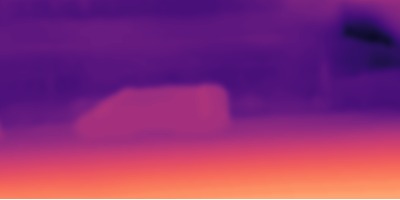}
& \hspace{-0.15cm} \includegraphics[width=1.98cm]{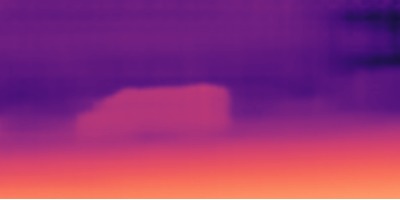}
& \hspace{-0.15cm} \includegraphics[width=1.98cm]{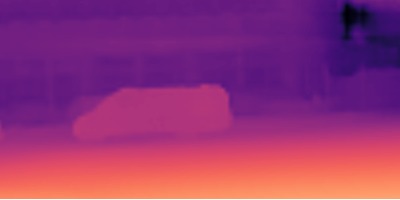} \\ 
\vspace{-0.05cm}
\hspace{-0.3cm} \includegraphics[width=1.98cm]{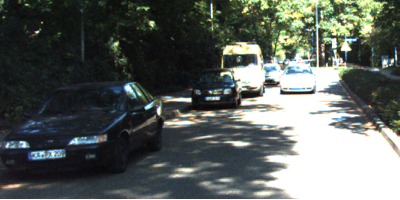} 
& \hspace{-0.15cm} \includegraphics[width=1.98cm]{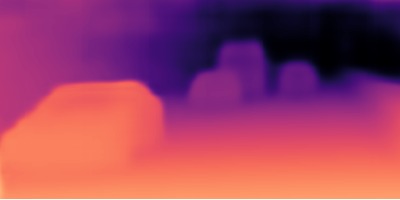}
& \hspace{-0.15cm} \includegraphics[width=1.98cm]{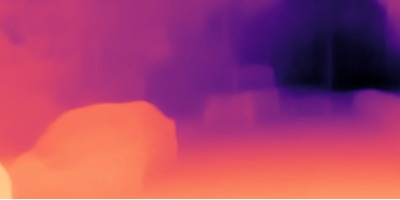}
& \hspace{-0.15cm} \includegraphics[width=1.98cm]{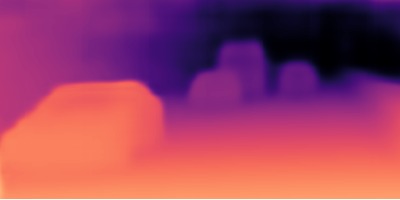}
& \hspace{-0.15cm} \includegraphics[width=1.98cm]{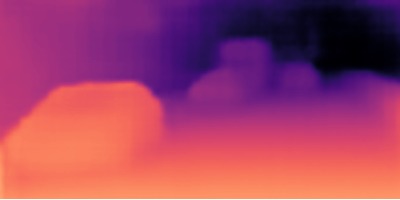}
& \hspace{-0.15cm} \includegraphics[width=1.98cm]{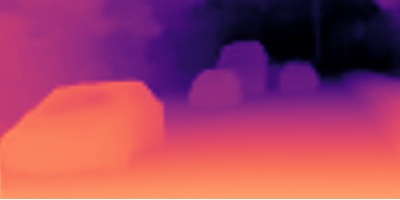} \\
\vspace{-0.05cm}
\hspace{-0.3cm} \includegraphics[width=1.98cm]{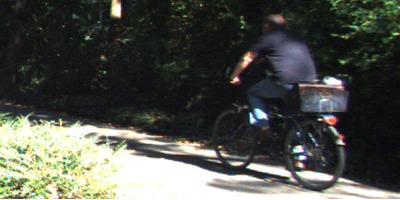} 
& \hspace{-0.15cm} \includegraphics[width=1.98cm]{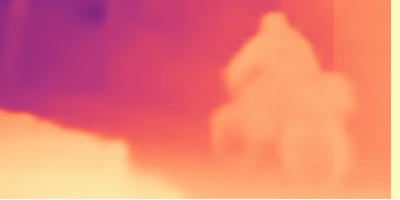}
& \hspace{-0.15cm} \includegraphics[width=1.98cm]{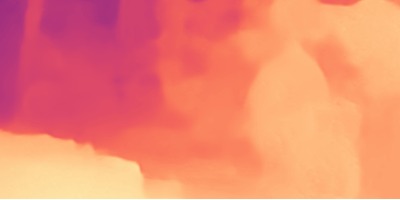}
& \hspace{-0.15cm} \includegraphics[width=1.98cm]{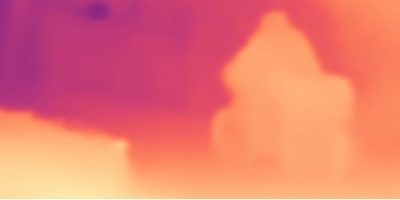}
& \hspace{-0.15cm} \includegraphics[width=1.98cm]{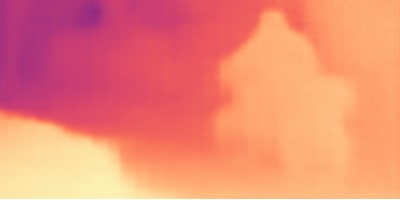}
& \hspace{-0.15cm} \includegraphics[width=1.98cm]{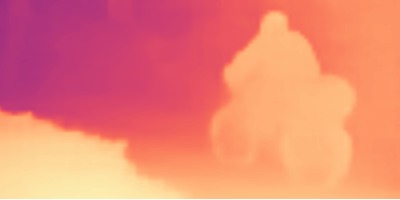} 
\end{tabular}$
\end{center}
\vspace{-20pt}
\caption{\small Qualitative results on KITTI. \ours~predicts more accurate depths, for instance, for building and car (first row), far-away cars and van (second row), and biker (third row).}
\label{fig:visualizations}
\vspace{-15pt}
\end{figure*}


\noindent\textbf{On KITTI.} 
The depth estimation accuracy results and comparison with existing methods are presented in Table~\ref{tab:KITTI}. Our proposed \ours~approach sets the new SOTA accuracy on KITTI, outperforming both latest monocular methods, e.g., iDisc~\cite{piccinelli2023idisc}, and video depth methods, e.g.,~\cite{yasarla2023mamo}. For ManyDepth~\cite{watson2021temporal} and TC-Depth~\cite{ruhkamp2021attention}, since the original models are trained in a self-supervised setting, we use the numbers from their fully-supervised versions retrained in~\cite{yasarla2023mamo}, referring them as ManyDepth-FS and TC-Depth-FS.

Fig.~\ref{fig:visualizations} shows a visual comparison of the depths predicted by \ours~ and existing SOTA methods. It can be seen that our predicted depth is more accurate and captures more details of the scene. For instance, sharp depth boundaries are predicted for the biker in the last example, even though this is a challenging case where the contrast between the biker and the background is low. 

\begin{figure}[t!]
\begin{center}$
\centering
\begin{tabular}{c c c c c}
 & \textbf{t=1} & \textbf{t=2} & \textbf{t=3} & \textbf{t=4}  \\
\vspace{-0.05cm} \hspace{-0.3cm} \rotatebox{90}{\quad \ \text{\textbf{Input}}} &  \includegraphics[width=2.2cm]{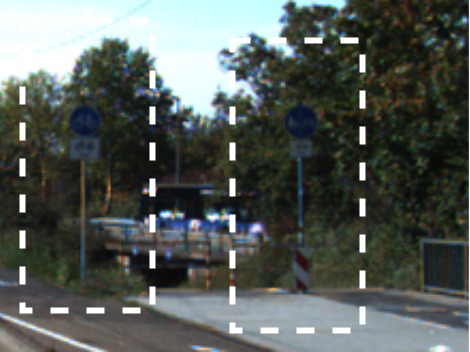} 
& \hspace{-0.15cm} \includegraphics[width=2.2cm]{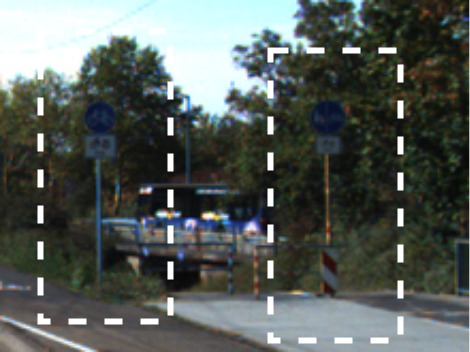} 
& \hspace{-0.15cm} \includegraphics[width=2.2cm]{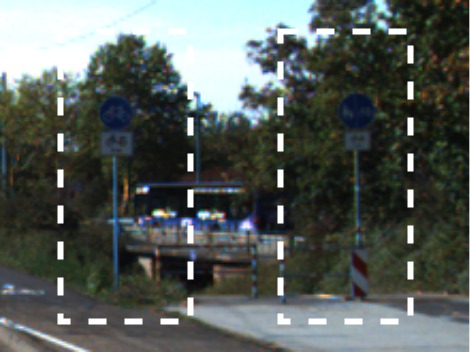} 
& \hspace{-0.15cm} \includegraphics[width=2.2cm]{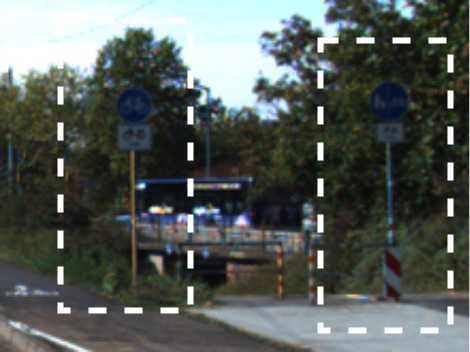} \\
\vspace{-0.05cm} \hspace{-0.3cm} \rotatebox{90}{\quad \ \text{\textbf{NVDS}}} & \includegraphics[width=2.2cm]{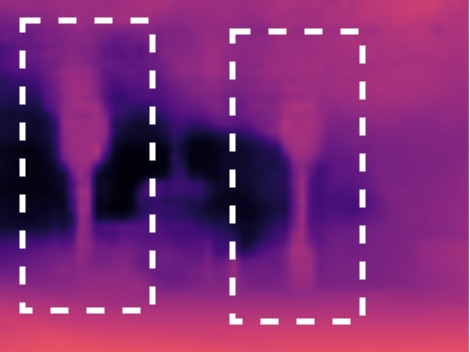} 
& \hspace{-0.15cm} \includegraphics[width=2.2cm]{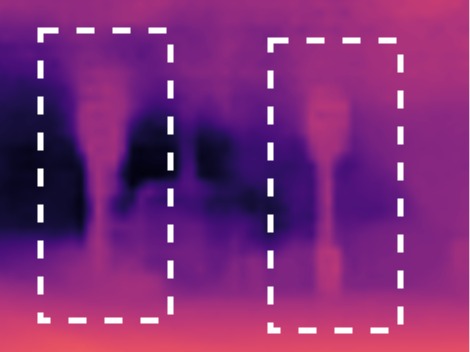} 
& \hspace{-0.15cm} \includegraphics[width=2.2cm]{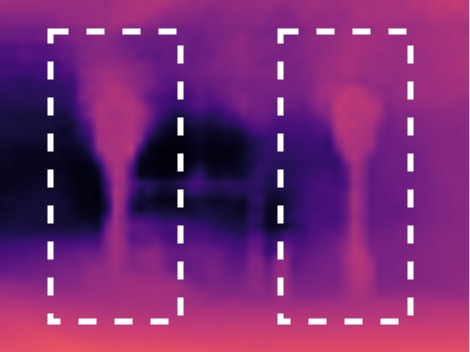} 
& \hspace{-0.15cm} \includegraphics[width=2.2cm]{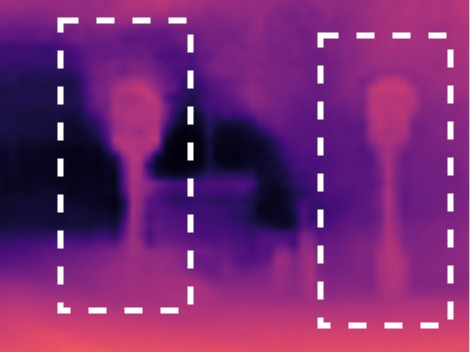} \\
\vspace{-0.05cm} \hspace{-0.3cm} \rotatebox{90}{\quad \ \text{\textbf{MAMo}}}  & \includegraphics[width=2.2cm]{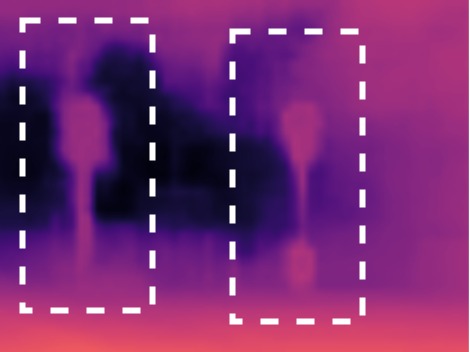} 
& \hspace{-0.15cm} \includegraphics[width=2.2cm]{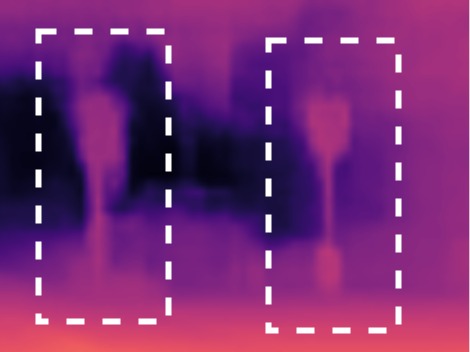} 
& \hspace{-0.15cm} \includegraphics[width=2.2cm]{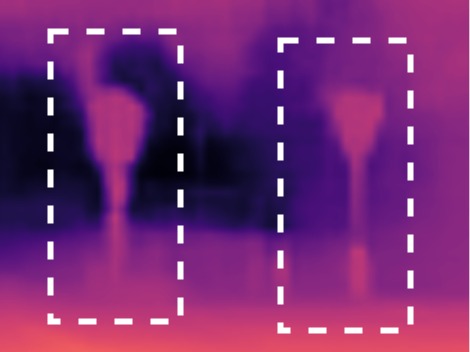} 
& \hspace{-0.15cm} \includegraphics[width=2.2cm]{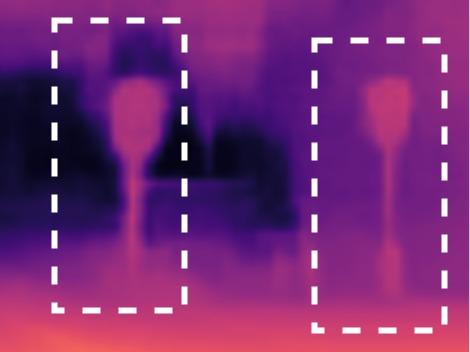} \\
\vspace{-0.05cm} \hspace{-0.3cm} \rotatebox{90}{\ \text{\textbf{\scriptsize{FutureDepth}}}} & \includegraphics[width=2.2cm]{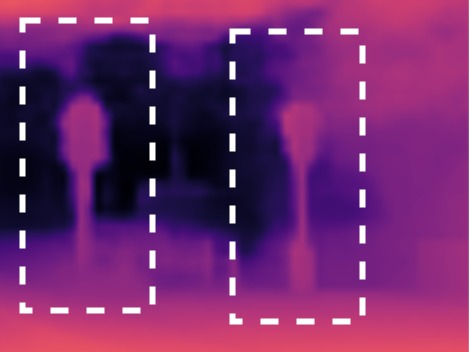} 
& \hspace{-0.15cm} \includegraphics[width=2.2cm]{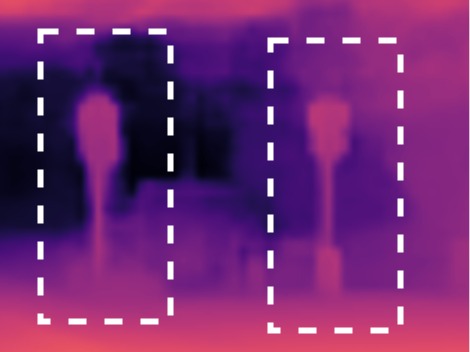} 
& \hspace{-0.15cm} \includegraphics[width=2.2cm]{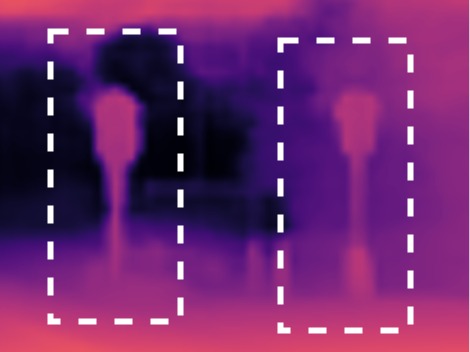} 
& \hspace{-0.15cm} \includegraphics[width=2.2cm]{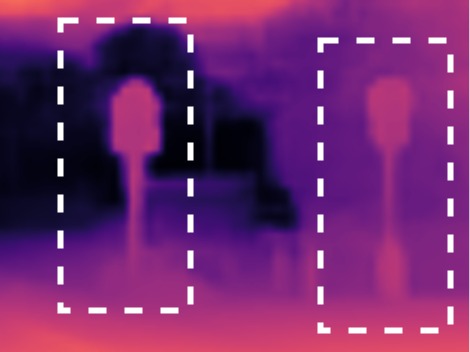} \\
\end{tabular}$
\end{center}
\vspace{-20pt}
	\caption{\small Sample patches from 4 consecutive frames. \ours~is more temporally consistent and accurate than existing SOTA.}
	\label{fig:time_seq}
	\vspace{-5pt}
\end{figure}

In addition, we evaluate the temporal consistency in Table~\ref{tab:TC_comp}. It can be seen that the temporal consistency of \ours~is significantly better than existing monocular and video methods. Fig.~\ref{fig:time_seq} shows a visual comparison of predicted depths on consecutive frames by \ours~and SOTA video depth estimation methods. We see that the depth predictions by MAMo and NVDS are inconsistent and noisy across frames, whereas our prediction is more temporally consistent and accurate.

We further measure the average model runtime. We see that our proposed \ours~is significantly more efficient as compared to existing video depth estimation models, and has comparable or better runtime as compared to SOTA monocular models.

\begin{table*}[t!]
    \caption{Temporal consistency and runtime on KITTI. All models use Swin-L as the encoder except for NVDS which uses DPT-L. Inference times are computed using NVIDIA RTX-3080 GPU with 11GB memory.}
	\label{tab:TC_comp}
    \centering
    \vspace{-8pt}
    \resizebox{0.82\linewidth}{!}{
    \normalsize
    \begin{tabular}{ll|ccc|c}
    \hline
    Type & Method & \quad rTC $\uparrow$ \quad & \quad aTC $\downarrow$ \quad & \quad OPW $\downarrow$ \quad & \quad  Runtime (ms) $\downarrow$ \quad
    \\
    \hline
    \multirow{3}{*}{SF} & NeWCRFs & 0.914 & 0.116 & 0.501 & 28 \\
    & iDisc & 0.923 & 0.108 & 0.486 & 61\\
    & GEDepth & 0.919 & 0.133 & 0.441 & 177\\
    \hline
    \multirow{4}{*}{MF} & Many-Depth-FS & 0.920 & 0.111 & 0.497 & 488 \\
    & TC-Depth-FS & 0.901 & 0.122 & 0.516 & 376 \\
    & NVDS & 0.951 & 0.096 & 0.356 & 930 \\
    & MAMo & 0.963 & 0.088 & 0.328 & 122 \\
    \hline
    \multirow{2}{*}{MF (ours)} & Baseline & 0.900 & 0.126 & 0.540 & 32 \\
    & FutureDepth & \textbf{0.988} & \textbf{0.076} & \textbf{0.281} & 49 \\

        \hline
    \end{tabular}
    }
    \vspace{-8pt}
    
\end{table*}


\begin{table*}[t!]
    \caption{\small Quantitative results on DDAD. $\uparrow$ ($\downarrow$) means higher (lower) is better. Best numbers are highlighted in bold. MF (SF) means multi-frame (single-frame) methods.}
	\label{tab:DDAD}
    \centering
    \vspace{-8pt}
    \resizebox{0.82\linewidth}{!}{
    \small
    \begin{tabular}{ll|cccc}
    \hline
    Type & {Method}                                
                &\multicolumn{1}{c}{\quad RMSE $\downarrow$ \quad } & \quad Sq Rel$\downarrow$ \quad &\multicolumn{1}{c}{\quad Abs Rel$\downarrow$ \quad} & \quad OPW$\downarrow$ \quad \\ \hline
                \multirow{3}{*}{SF}  &NeWCRFs~\cite{yuan2022newcrfs}     & {10.98}   & {2.831}   & {0.291}    &   0.622   \\ 
                 & iDisc~\cite{piccinelli2023idisc}      & {8.99}     & {1.854} & {0.163}    &   0.596  \\ 
                 & GEDepth~\cite{yang2023gedepth}  & {10.60}  & {2.119}  & {0.157}   &   0.571    \\ \hline
                 \multirow{4}{*}{MF}  &Many-Depth-FS~\cite{watson2021temporal}     & {12.35}   & {3.946}   & {0.292}    &   0.544   \\ 
                 & TC-Depth-FS~\cite{ruhkamp2021attention}        & {12.11}     & {3.788} & {0.283}    &   0.699   \\ 
                 & NVDS~\cite{wang2023neural}        & {9.24}   & {1.995}   & {0.174}    &   0.496   \\ 
                 & MAMo~\cite{yasarla2023mamo}       & {8.45}  & {1.772}  & {0.150}   &   0.464     \\ \hline
                 \multirow{2}{*}{MF (ours)} & Baseline        & {11.36}     & 3.216 & 0.232 & 0.681 \\ 
                 & \ours      & {\textbf{7.72}}    & \textbf{1.290}  & {\textbf{0.114}}    &   \textbf{0.366}   \\ \hline
    \end{tabular}
    }
    \vspace{-8pt}
\end{table*}

\noindent\textbf{On DDAD.} Table~\ref{tab:DDAD} shows the depth prediction results of models trained and evaluated on DDAD. All methods are trained using Swin-L or DPT-L encoder for fair comparison. It can be seen that our proposed \ours~outperforms the existing SOTA methods. Additionally, we perform zero-shot testing by evaluating KITTI-trained models on DDAD and observe that \ours~has better generalizability as compared to existing models. The zero-shot testing results are included in the supplementary.

\noindent\textbf{On Sintel.} Table~\ref{tab:NYUv2}~(right) shows zero-shot evaluation results on Sintel. We see that \ours~significantly outperforms existing SOTA video depth estimation models, with better accuracy and temporal consistency. This demonstrates the strong generalization ability of our proposed \ours~approach. 

\vspace{-10pt}
\subsection{Ablation Study}
\label{sec:ablation_study}
\vspace{-8pt}
We conduct comprehensive experiments to investigate different components in our proposed \ours~ pipeline. Table~\ref{tab:Abl_kitti} summarizes the results. 
We assess two baselines: (1) single-frame (SF) baseline that operates on single input frames and uses the same encoder and decoder architecture as our main model, with channel numbers modified accordingly for the single-frame setting; (2) multi-frame (MF) baseline that is described in Sec.~\ref{sec:exp_setup}. 
We then incrementally add our proposed modules to evaluate their effects, including \PNet, R-Net with random masking and adaptive masking, and refinement network.

\begin{figure}[t!]
	\centering
	\includegraphics[width=\linewidth]{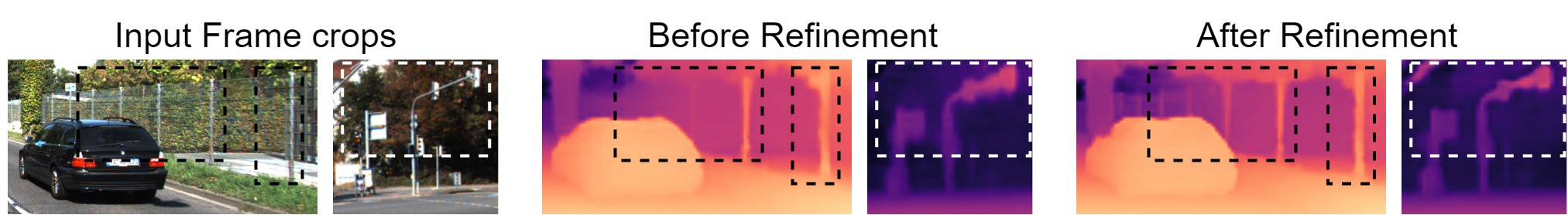} 
	\vskip -8pt 
	\caption{\small Depth estimation quality is considerably improved after refinement, e.g., fence (first sample), traffic light (second sample).}
	\label{fig:rebuttal_refine}
	\vspace{-8pt}
\end{figure}

We can see that by naively extending a single-frame model to a multi-frame model only brings minimal gains; see the first two rows in Table~\ref{tab:Abl_kitti}. Using \PNet~significantly improves the baseline in terms of both accuracy (e.g., $16\%$ in Sq Rel) and temporal consistency (e.g., $42\%$ in OPW). In addition, R-Net provides visible improvements on top of the baseline, and together with adaptive masking brings more performance gains. Using both \PNet~and R-Net jointly generates significant improvements as compared to using either of them alone. Finally, using the refinement network can further reduce prediction errors. Fig.~\ref{fig:rebuttal_refine} provides visual examples of how the refinement network improves the details of the predicted depth maps.

\begin{table*}[t!]
    \caption{\small Ablation study on KITTI. We use Swin-L encoder for all variants. $\uparrow$ ($\downarrow$) means higher (lower) is better. AM means adaptive masking. RM means random masking. 
    }
	\label{tab:Abl_kitti}
    \centering
    \vspace{-8pt}
    \resizebox{1\linewidth}{!}{
    \small
    \begin{tabular}{l|c|cccc|cccc}
    \hline
    \quad Model \quad & \quad Type \quad & \quad R-Net \quad & \quad AM \quad & \quad F-Net \quad & \quad Refine \quad & \quad Sq Rel$\downarrow$ \quad & \quad RMSE$\downarrow$ \quad & \quad $\delta<1.25\uparrow$ \quad & \quad OPW$\downarrow$ \quad \\ \hline
    Baseline & SF & & & & & 0.156 & 2.098 & 0.974 & 0.544 \\
    \hline
    \multirow{5}{*}{Baseline} & \multirow{5}{*}{MF} & & & & &  0.154 & 2.094  & 0.975 & 0.540 \\
    & & & & $\checkmark$ &   & 0.129 & 1.978 & 0.981 & 0.311  \\ 
    & & $\checkmark$ (RM) &  &  &   & 0.148 & 2.040  &  0.976 & 0.478 \\
    & & $\checkmark$ & $\checkmark$ & & & 0.136 & 1.999  & 0.980 & 0.416  \\
    & & $\checkmark$ & $\checkmark$ & $\checkmark$ &   & 0.122 & 1.931  & \textbf{0.983} & 0.284  \\ 
    \hline
    \ours & MF & $\checkmark$ & $\checkmark$ & $\checkmark$ & $\checkmark$  & \textbf{0.119} & \textbf{1.920}  & \textbf{0.983} & \textbf{0.281}  \\ 
        \hline
    \end{tabular}
    }
    \vspace{-5pt}
\end{table*}


\vspace{-8pt}
\section{Related work}
\label{sec:related}

\noindent\textbf{Monocular Depth Estimation (MDE).}  
MDE is the task of estimating depth based on a single image. Early approaches utilize conventional or hand-crafted features~\cite{michels2005high, nagai2002hmm, saxena2005learning, saxena2008make3d, wang2015depth}. More recently, deep-learning-based methods have shown significant improvements. One general approach treats depth estimation as a continuous regression task~\cite{monodepth17,yuan2022newcrfs,zhu1232020mda,cai2021x,shi2023ega}. Other works, e.g., \cite{bhat2021adabins,li2022binsformer,fu2018deep}, consider depth prediction as a classification or ordinal regression task, and put the depth values of a scene into discrete bins. While there has been extensive research on MDE, this approach inherently ignores the temporal information in video data, which is usually available in practical applications.

\noindent\textbf{Video Depth Estimation (VDE).}
Recently, researchers have looked into utilizing  multiple frames for depth estimation in a deep learning framework. \cite{watson2021temporal} introduces a cost-volume-based method to leverage consecutive frames for depth estimation, which is further extended by~\cite{Long_2021_CVPR} and~\cite{sayed2022simplerecon} to include additional frames through cost-volume aggregation. 
Cost volume architectures, however, incur high computation and memory costs, making them challenging to run resource-constrained platforms and extend to more frames. 
Other works explore the use of recurrent neural networks, but only obtain sub-optimal accuracy~\cite{eom2019temporally, zhang2019exploiting, patil2020don}. 
Recently, researchers have started to adopt attentions in video depth estimation. Early attempts do not achieve SOTA performance, even when compared to the latest MDE models~\cite{cao2021learning, wang2022less}. 
Some researchers leverage optical flow in video depth estimation~\cite{eom2019temporally,xie2020video,yasarla2023mamo} and the latest of them~\cite{yasarla2023mamo} achieves significantly better accuracy.  
However, in addition to requiring optical flow estimation, \cite{yasarla2023mamo} requires backpropagation-based feature updates on the fly, which is computationally expensive. Another latest attention-based work~\cite{wang2023neural} achieves SOTA accuracy, but also incurs significant computational costs. Another line of work focuses on test-time training~\cite{luo2020consistent, kopf2021robust}. While they can achieve temporally consistent depths by overfitting to one test video, such training-based approaches are too slow for real-time applications, computationally infeasible for resource-constrained devices, and not generalizable.

\vspace{-8pt}
\section{Conclusion}
\vspace{-8pt}
In this paper, we proposed a novel and efficient video depth estimation method, \ours. Specifically, we proposed a Future Prediction Network (\PNet), which is trained using iterative future feature prediction, to capture key motion cues to enhance depth prediction. We also proposed a Reconstruction Network (R-Net), which is trained via masked auto-encoding on multi-frame features with a learnable masking. In this way, R-Net learns to identify multi-frame correspondences, which benefits depth estimation. At inference, \PNet~and R-Net generates query features containing key motion and scene information, which is incorporated into the depth decoding process through cross-attention. Moreover, we use these queries in a refinement stage to further improve accuracy. Extensive results on benchmark datasets such as NYUDv2, KITTI, DDAD, and Sintel, have demonstrated the efficacy of our proposed \ours~method. \ours~sets the new state-of-the-art accuracy and at the same time, is more efficient than latest video and monocular depth models.

\noindent\textbf{Limitations.} There are a few aspects of video depth estimation not addressed in this work, which can be interesting for future research. For instance, we do not propose specific treatment for cases where an object becomes occluded and then re-appears in a set of consecutive frames. Proper handling of occlusion can lead to better motion and correspondence understanding and as a result, more accurate depth estimation.


\section*{Acknowledge}
\vspace{-10pt}
We would like to thank Hanno Ackermann for the insightful feedback and discussion.
\bibliographystyle{splncs04}
\bibliography{main}
\clearpage
\setcounter{page}{1}
\appendix

\section{Additional Ablation Studies}
\noindent\textbf{Sampling Techniques.} We perform experiment to ablate different sampling techniques used in reconstruction network (R-Net) on KITTI dataset. In Table~\ref{tab:Abl_samp}, we can see that the proposed 
adaptive sampling technique benefits the R-Net in learning better spatial and temporal representation of the feature volume $\hat{V}_{1,T}$ for given sequence of frames $I_{1,T}$. In Fig.~\ref{fig:example-mask} of the main paper, we can see adaptive sampling generates masks that preserve representations of important object information and exclude unnecessary back ground information, thus benefiting \ours~depth estimation performance. 
Note that in this experiment we do not use \PNet~and refinement network in \ours, \textit{i.e.,} we perform experiments using Baseline (MF) + R-Net to understand the effect of sampling techniques more clearly. 

\begin{table}[h!]
    \caption{Using different masking techniques on KITTI (Eigen split) dataset. We perform this experiment using Swin-L for \ours~encoder.  Here we set $T=4$.}
	\label{tab:Abl_samp}
    \centering
    \vspace{-6pt}
    \normalsize
    \begin{tabular}{l|ccccc}
    \hline
        Masking Method & Abs Rel$\downarrow$  & Sq Rel$\downarrow$ & RMSE$\downarrow$  & $\text{RMSE}_{log}\downarrow$ & $\delta<1.25\uparrow$ \\ \hline
         Random & 0.051  & 0.148 & 2.040  & 0.077 &  0.976   \\ 
         Tube  & 0.050 & 0.146 & 2.035 & 0.077 & 0.977 \\
         Adaptive (ours)  & 0.048 & 0.136 & 1.999 & 0.073 & 0.980 \\
        \hline
    \end{tabular}
    \vspace{-0em}
\end{table}

\noindent\textbf{Number of iteration ($L$) in \PNet.} We perform experiment on KITTI dataset to ablate on the number of iterations ($L$) in the future prediction network (\PNet), which shows the impact of $Q_{motion,1,T}$ motion queries on \ours~performance. As shown in Table~\ref{tab:Abl_p_iter}, as the iterations in \PNet~increases,  \PNet~can predict the current motion  and future of objects over the time frames from 1 to $L+T$ and generate beneficial $Q_{motion,1,T}$ queries that provide temporal information to \ours~decoder and benefits \ours~performance. Note, here we didn't use the refinement network in \ours~for this experiment.

\begin{table}[h!]
    \caption{Ablation study for different number of iterations ($L$) in future prediction on KITTI (Eigen split) dataset. We perform this experiment using Swin-L for \ours~encoder.  Here we set  $T=4$.}
	\label{tab:Abl_p_iter}
    \centering
    \vspace{-6pt}
    \small
    \begin{tabular}{l|ccccc}
    \hline
        Metric & 0  & 1 &  2 & 3 &  4 \\ \hline
         RMSE$\downarrow$ & 1.999  & 1.976 & 1.962  & 1.944 &  1.931   \\ 
         Sq Rel$\downarrow$  & 0.136 & 0.130 & 0.136 & 0.124 & 0.122 \\
        \hline
    \end{tabular}
    \vspace{-0em}
\end{table}

\noindent\textbf{Number of refinement steps ($N$).} We perform experiment on KITTI dataset to ablate on the number of refinement steps ($N$), in order to analyze the impact of refinement network using $Q_{scene,1,T}$ and $Q_{motion,1,T}$ queries on \ours~performance. The result is shown in Table~\ref{tab:Abl_refine}.

\begin{table}[h!]
    \caption{Ablation study for different of refinement steps ($N$) on KITTI (Eigen split) dataset. We perform this experiment using Swin-L for \ours~encoder. Here we set $L=4$ and $T=4$.}
	\label{tab:Abl_refine}
    \centering
    \vspace{-6pt}
    \small
    \begin{tabular}{l|cccc}
    \hline
        Metric & 0  & 1 &  2 & 3 \\ \hline
         RMSE$\downarrow$ & 1.931  & 1.924 & 1.922  & 1.920   \\ 
         Sq Rel$\downarrow$  & 0.122 & 0.120 & 0.119 & 0.119 \\
        \hline
    \end{tabular}
    \vspace{-0em}
\end{table}

\noindent\textbf{Number of frames ($T$).} We perform experiment on KITTI dataset to ablate on the number of frames $T$ in a video/multi-frame sequence. Here, Table~\ref{tab:Abl_T} shows the performance of \ours~when processing $T$ video frames in a batch simultaneously. 
\begin{table}[h!]
    \caption{Ablation study on different of numbers of frames ($T$) in video sequence on KITTI (Eigen split) dataset. We perform this experiment using Swin-L for \ours~encoder. Here we set $L=4$.}
    \label{tab:Abl_T}
    \centering
    \vspace{-6pt}
    \small
    \begin{tabular}{l|cccc}
    \hline
        Metric & 3  & 4 &  6 & 8 \\ \hline
         RMSE$\downarrow$ & 1.932  & 1.920 & 1.906  & 1.911   \\ 
         Sq Rel$\downarrow$  & 0.122 & 0.119 & 0.114 & 0.116 \\
        \hline
    \end{tabular}
    \vspace{-0em}
\end{table}

\subsection{What happens when the adaptive sampler is used during inference?}
We perform experiment on KITTI dataset to study the benefits of adaptive sampler during inference time. Note that we train \ours~with adaptive sampler where we choose the masking ratio from $r \in [0.6,0.9]$ to train R-Net. Here, we set $L=4$, $T=4$, and $N=3$ for \ours~training and inference. During inference we ablate on different values of masking ratio ($r=0,0.2,0.4,0.6,0.9$) for adaptive sampler in R-Net. Table~\ref{tab:Abl_adpt} demonstrates that the inference-time utilization of adaptive sampling in R-Net can benefit \ours~by assigning significance to critical attentions and enhancing the feature volume ($\hat{V}_{1,T}$), which is subsequently processed by the \ours~decoder.

\begin{table}[h!]
    \caption{Ablation study on different masking ratios of adaptive sampler during inference on KITTI (Eigen split) dataset. We perform this experiment using Swin-L for \ours~encoder. Here we set $L=4$, $T=4$ and $N=3$.}
	\label{tab:Abl_adpt}
    \centering
    \vspace{-6pt}
    \normalsize
    \begin{tabular}{l|ccccc}
    \hline
        Metric & 0.0  & 0.2 &  0.4 & 0.6 & 0.9 \\ \hline
         RMSE$\downarrow$ & 1.920  &  1.906  & 1.892 & 1.911 & 1.956 \\ 
         Sq Rel$\downarrow$  & 0.119 & 0.114 & 0.108 & 0.111 & 0.133 \\
        \hline
    \end{tabular}
    \vspace{-0em}
\end{table}

\newpage
\section{Zero-Shot Evaluation} 
In Table~\ref{tab:DDAD_sup}, we assess the performance of KITTI-trained models on DDAD to evaluate their generalization capabilities. The results indicate that our proposed \ours~surpasses existing state-of-the-art methods. The evaluation of KITTI-trained models on DDAD demonstrates that \ours~exhibits superior generalizability compared to other models.
\begin{table}[h!]
    \caption{Quantitative results on DDAD dataset for distances up to 150 meters. The input frame resolution is $1216\times 1936$.}
    \label{tab:DDAD_sup}
    \centering
    \vspace{-5pt}
    \small
    \begin{tabular}{l|l|ccc}
    \hline
        Method & Encoder & Sq Rel$\downarrow$ & RMSE$\downarrow$ & $\delta<1.25\uparrow$ \\ \hline
       ManyDepth-FS\cite{watson2021temporal} & ResNet50 & 5.471& 16.123  & 0.744 \\ 
        ManyDepth-FS\cite{watson2021temporal} & Swin-L & 4.211 & 13.899 & 0.784 \\ 
        TC-Depth-FS\cite{ruhkamp2021attention} & ResNet50 & 5.285 & 15.121 & 0.777 \\ 
        AdaBins\cite{bhat2021adabins} & \cite{tan2019efficientnet}  & 4.950 &  15.228	&  0.780 \\
        AdaBins\cite{agarwal2022depthformer} & \cite{xie2021segformer} & 4.791	& 14.595	& 0.789	\\
        NeWCRFs\cite{yuan2022newcrfs} & Swin-L & 4.041 & 11.956 & 0.816 \\ 
        PixelFormer\cite{Agarwal_2023_WACV} &  Swin-L & 4.474& 12.467 & 0.802 \\ 
         MAMo\cite{yasarla2023mamo} & Swin-L & 3.349 & 11.094 & 0.870   \\  \hline
        Baseline (ours) & Swin-L & 4.506 & 12.841 & 0.804   \\
        \ours~ (ours) & Swin-L & 2.960 & 10.016 & 0.833   \\
        \hline
    \end{tabular}
    \vspace{-0.0em}
\end{table}

\newpage
\section{Training and Infernece Algorithms}
Algorithm~\ref{alg:pre_train} outlines the steps involved in the pre-training phase of \ours, while Algorithm~\ref{alg:train} details the steps of the main training phase of \ours. Additionally, Algorithm~\ref{alg:inference} provides the steps for evaluating \ours~during inference.

\begin{algorithm}[h!]
\caption{Pretraining of \ours}\label{alg:pre_train}
\footnotesize{
\begin{algorithmic}[]
\STATE \textbf{Input}: Train dataset $\mathcal{D}$ which consists of training videos and corresponding ground truth depths. Training video or sequence of frames, $I_{1,T} =\{I_1,...,I_T\}$ and $D^{gt}_{1,T} =\{D^{gt}_0,...,D^{gt}_T\}$
\STATE \textbf{Model}: $en(\cdot)$: encoder of \ours, $de(\cdot)$: decoder of \ours, $g(\cdot)$: reconstruction network and $h(\cdot)$: future prediction network, $rf(\cdot)$: refinement network, $s(\cdot)$: adaptive sampler\\
\#\#\#\#\#\#  pretraining $en(\cdot)$, $de(\cdot)$ \#\#\#\#\#\#
\FOR{epoch = 1$\rightarrow$5}
    \FOR {$I_{1,T},D^{gt}_{1,T} \in \mathcal{D}$}
        \STATE ${D}_{1,T}\: =\: de(en(I_{1,T}))$
        \STATE \textit{SILogLoss} (${D}_{1,T}$, ${D}^{gt}_{1,T}$)
        \STATE \textit{Update parameters of} $en(\cdot)$, $de(\cdot)$
    \ENDFOR
\ENDFOR   \\
\#\#\#\#\#\# pretraining reconstruction network $g(\cdot)$\#\#\#\#\#\# \\
\FOR{epoch = 1$\rightarrow$3}
    \STATE freeze $en(\cdot)$ weights
    \FOR {$I_{1,T},D^{gt}_{1,T} \in \mathcal{D}$}
        \STATE $V_{1,T}\: =\: en(I_{1,T})$
        \STATE generate random mask $M_{1,T}$
        \STATE $\hat{V}_{1,T}\: =\: g(M_{1,T}\odot V_{1,T})$
        \STATE L2-loss between $V_{1,T}$ and $\hat{V}_{1,T}$
        \STATE \textit{Update parameters of} $g(\cdot)$
    \ENDFOR
\ENDFOR  
\\
\end{algorithmic}
}
\end{algorithm}

\begin{algorithm}[h!]
\caption{Main Training of \ours}\label{alg:train}
\footnotesize{
\begin{algorithmic}[]
\STATE \textbf{Input}: Train dataset $\mathcal{D}$ which consists of training videos and corresponding ground truth depths. Training video or sequence of frames, $I_{1,T} =\{I_1,...,I_T\}$ and $D^{gt}_{1,T} =\{D^{gt}_0,...,D^{gt}_T\}$
\STATE \textbf{Model}: $en(\cdot)$: encoder of \ours, $de(\cdot)$: decoder of \ours, $g(\cdot)$: reconstruction network and $h(\cdot)$: future prediction network, $rf(\cdot)$: refinement network, $s(\cdot)$: adaptive sampler\\
\#\#\#\#\#\# Training \ours~network \#\#\#\#\#\# \\
initialize weights of $h(\cdot)$ with $g(\cdot)$
\FOR{every epoch}
    \FOR {$I_{1,T},D^{gt}_{1,T} \in \mathcal{D}$}
        \STATE --------------------------------------------------------------------
        \STATE \#\#\# step-1 updating $h(\cdot)$, $s(\cdot)$, $g(\cdot)$ weights\#\#\#
        \STATE freeze $en(\cdot)$, $de(\cdot)$  weights
        \STATE $V_{1,T}\: =\: en(I_{1,T})$; \quad $M_{1,T}\: =\: s(V_{1,T})$
        \STATE $\tilde{V}_{1,T}\: =\: V_{1,T}$
        \FOR{i=1$\rightarrow$ L}
            \STATE \quad $\tilde{V}_{i+1,i+T}\: =\: h(\tilde{V}_{i,i+T-1})$
        \ENDFOR
        \STATE $\hat{V}_{1,T},\:Q_{scene,1,T} \: =\: g(M_{1,T}\odot V_{1,T})$
        \STATE  $Q_{all,1,T} = \text{cross-attn(}Q_{scene,1,T}, Q_{motion,1,T})$ 
        \STATE ${D}_{1,T}\: =\: de(\hat{V}_{1,T},Q_{all,1,T})$
        \STATE compute loss $\mathcal{L}_{F}$ in Eq. 1 (main paper)
        \STATE compute loss $\mathcal{L}_{A}$ (refer section 2.2)
        \STATE compute loss $\mathcal{L}_{R}$ in Eq. 2 (main paper)
        \STATE \textit{Update parameters of} $h(\cdot)$, $s(\cdot)$, $g(\cdot)$
        \STATE --------------------------------------------------------------------
        \STATE \#\#\# step-2 updating \ours's $en(\cdot)$, $de(\cdot)$, $rf(\cdot)$ weights\#\#\#
        \STATE freeze $s(\cdot)$, $g(\cdot)$, $h(\cdot)$   weights
        \STATE $V_{1,T}\: =\: en(I_{1,T})$\; 
        \STATE $\hat{V}_{1,T},\:Q_{scene,1,T} \: =\: g( V_{1,T})$
        \STATE get $Q_{motion,1,T}$ from future prediction \PNet~$h(\cdot)$
        \STATE  $Q_{all,1,T} = \text{cross-attn(}Q_{scene,1,T}, Q_{motion,1,T})$ 
        \STATE ${D}^0_{1,T}\: =\: de(\hat{V}_{1,T},Q_{all,1,T})$
        \FOR{i=1$\rightarrow$ N}
            \STATE \quad ${D}^i_{1,T}\: =\: rf({D}^{i-1}_{1,T},Q_{all,1,T})$
        \ENDFOR
        \STATE compute loss $\mathcal{L}_{D,final}$ in Eq. 3 (main paper)
        \STATE \textit{Update parameters of} \ours~'s $en(\dot)$, $de(\cdot)$, $rf(\cdot)$ weights
    \ENDFOR
\ENDFOR
\end{algorithmic}
}
\end{algorithm}

\begin{algorithm}[h!]
\caption{Inference of \ours}\label{alg:inference}
\footnotesize{
\begin{algorithmic}[]
\STATE \textbf{Input}: Test dataset $\mathcal{D}^{test}$ which consists of inference videos. Inference video or sequence of frames, $I_{1,T} =\{I_1,...,I_T\}$ 
\STATE \textbf{Model}: $en(\cdot)$: encoder of \ours~, $de(\cdot)$: decoder of \ours~, $g(\cdot)$: reconstruction network and $h(\cdot)$: future prediction network, $rf(\cdot)$: refinement network, $s(\cdot)$: adaptive sampler\\
\#\#\#\#\#\#Inference \ours~ network \#\#\#\#\#\# \\
\FOR {$I_{1,T} \in \mathcal{D}^{test}$}
    \STATE $V_{1,T}\: =\: en(I_{1,T})$
    \STATE $\hat{V}_{1,T},\:Q_{scene,1,T} \: =\: g( V_{1,T})$
    \STATE get $Q_{motion,1,T}$ from future prediction \PNet~ $h(\cdot)$
    \STATE  $Q_{all,1,T} = \text{cross-attn(}Q_{scene,1,T}, Q_{motion,1,T})$ 
    \STATE ${D}^0_{1,T}\: =\: de(\hat{V}_{1,T},Q_{all,1,T})$
    \FOR{i=1$\rightarrow$ N}
        \STATE \quad ${D}^i_{1,T}\: =\: rf({D}^{i-1}_{1,T},Q_{all,1,T})$
    \ENDFOR
    \STATE $D_{1,T} =\hat{D}^N_{1,T}$
\ENDFOR
\end{algorithmic}
}
\end{algorithm}

\newpage
\clearpage

\section{Details on \ours~Networks}

\begin{figure*}[t!]
    \centering
    \includegraphics[width=0.98\linewidth]{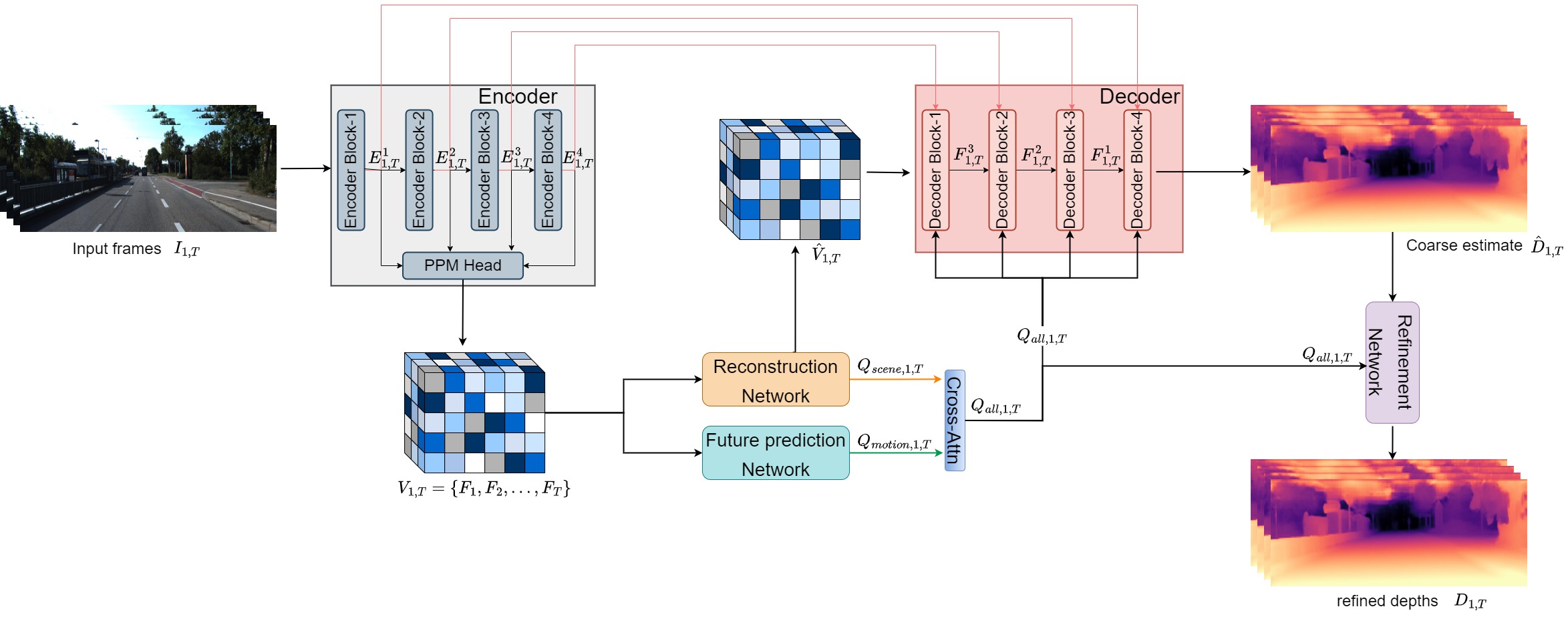} 
    \vskip -5pt 
    \caption{\small Our proposed \ours~ method.}
    \label{fig:RAP-detail}
    \vspace{-5pt}
\end{figure*}

\begin{figure}[t!]
    \centering
    \includegraphics[height=7cm]{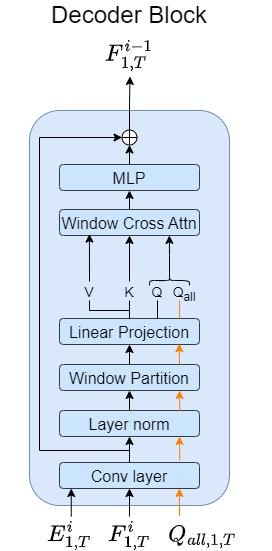} 
    \vskip -5pt 
    \caption{\small Skip attention module used as a building block for each decoder layer in \ours~.}
    \label{fig:SAM-deocder}
    \vspace{-5pt}
\end{figure}

\begin{figure*}[t!]
    \centering
    \includegraphics[width=0.99\linewidth]{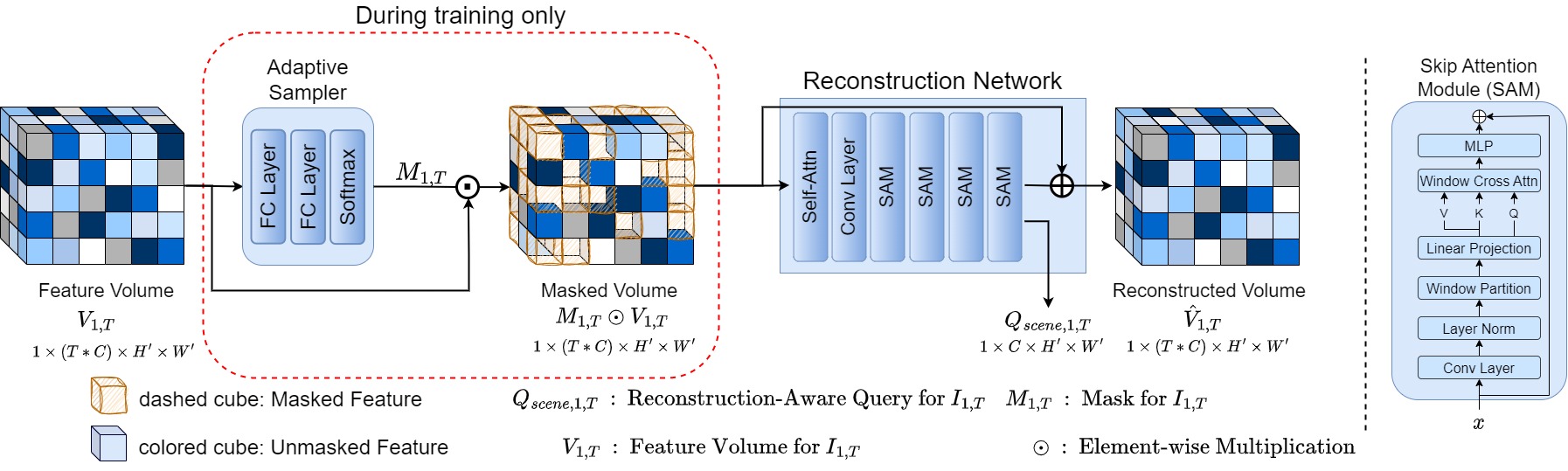} 
    \vskip -5pt 
    \caption{\small Reconstruction network (R-Net) in our proposed \ours~ framework. 
    }
    \label{fig:recon}
    \vspace{-1.5em}
\end{figure*}

\begin{figure}[t!]
    \centering
    \includegraphics[width=\linewidth]{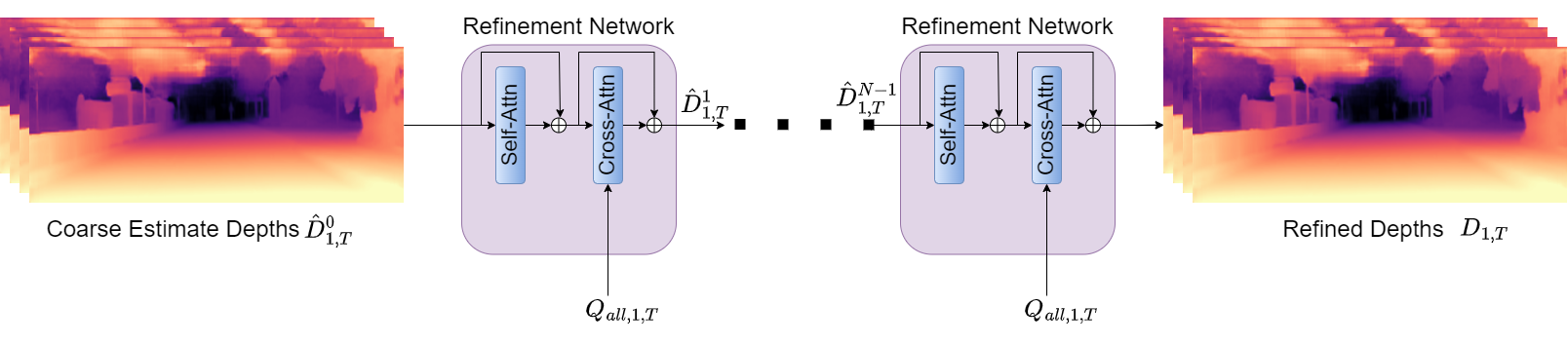} 
    \vskip -5pt 
    \caption{Overview of refinement network.}
    \label{fig:refine}
    \vspace{-1.0em}
\end{figure}

Fig.~\ref{fig:RAP-detail} shows the detailed network architecture of \ours.  PPM head used in \ours~encoder is similar to \cite{yuan2022newcrfs}. Note, cross-attention and self-attention layers used in \ours~are similar to \cite{wang2020linformer}. For example we use \cite{wang2020linformer} cross-attention layer to perform cross-attention between $Q_{scene}$ and $Q_{motion}$ to produce $Q_{all}$. Fig.~\ref{fig:SAM-deocder} shows the decoder block used in \ours~decoder. Fig.~\ref{fig:SAM-deocder} clearly show $Q_{all}$ queries which contains critical scenes and temporal cues are utilized by \ours~decoder. Fig.~\ref{fig:recon} shows overview architecture of R-Net. The mask generator consists of two fully-connected layers and a softmax layer. Based on the softmax scores, we keep the top $r\times P$ patches and mask out the rest, where $r$ is the masking ratio and $P$ is total number of patches. 
Fig~\ref{fig:refine} shows the overview of refinement network, where we utilize $Q_{all}$ and improve the coarse estimate depths predicted by \ours~decoder.  Refinement network contains a self-attention layer and a cross-attention layers as shown in Fig.~\ref{fig:refine}, where we first perform self-attention between the initial coarse depth predictions ${D}^0_{1,T}$, which are further cross-attended with the $Q_{all}$ to obtain ${D}^1_{1,T}$. We perform this progressive refinement process for $N$ steps to obtain final depth prediction ${D}_{1,T}$ (=$D^N_{1,T}$).


\newpage
\clearpage
\section{Details on Evaluation Metrics} 
We follow \cite{eigen2014depth} to use the following metrics to evaluate the performance of predicted depth outputs of different methods,
\begin{equation}\label{eqn:metrics}
\begin{array}{l}
\text {Abs Relative: } \frac{1}{\sum(K_t==1)} \sum_{k_t \in K, d_t \in D_t}k_t\|\frac{d_t - {d}_{t}^{gt}}{d_t^{gt}}\| \\
\text {Squared Relative: } \frac{1}{\sum(K_t==1)} \sum_{k_t \in K, d_t \in D_t} \frac{\left\|d_t-d_t^{gt}\right\|^2 }{ d_t^{gt}}  \\
\operatorname{RMSE}\left(\text{linear}\right): \sqrt{\frac{1}{\sum(K_t==1)} \sum_{k_t \in K, d_t \in D_t} \left\|d_t-d_t^{gt}\right\|^2}\\
\operatorname{RMSE~(log):~} \sqrt{\frac{1}{\sum(K_t==1)} \sum_{k_t \in K, d_t \in D_t}\left\|\log d_t-\log d_t^{gt}\right\|^2} \\
\delta < \text{thr: }  \frac{1}{\sum(K_t==1)}K_t\left[\text{Max}\left(\frac{D_t}{{D}_{t}^{gt}}, \frac{{D}_{t}^{gt}}{D_t}  \right)<\text{thr}\right]\\
\end{array}
\end{equation}
where $K_t$ is a depth validity mask, $D_t$ is predicted depth for image $I_t$ and $D_t^{gt}$ is ground-truth depth.

For evaluating temporal consistency, \cite{li2021enforcing} introduces the following metrics, 
\begin{equation}\label{eqn:temporal_eqn}
\begin{aligned}
    aTC_t &= \frac{1}{\sum(K_t==1)}K_t\|\frac{D_t - {D}_{t}^w}{D_t}\|,\\
    rTC_t &=  \frac{1}{\sum(K_t==1)}K_t\left[\text{Max}\left(\frac{D_t}{{D}_{t}^w}, \frac{{D}_{t}^w}{D_t}  \right)<\text{thr}\right],
\end{aligned}
\end{equation}
where $K_t$ is a depth validity mask, $D_t$ is predicted depth for $I_t$ and ${D}_{t}^w$ is warped from ${D}_{t-1}$ using optical flow. Following the protocol introduced by \cite{yasarla2023mamo} we use the optical flow generated by the latest SOTA FlowFormer~\cite{huang2022flowformer}. Optical flow based warping metric (OPW) is introduced by \cite{wang2022less},

\begin{equation}
\begin{gathered}
O P W_t=\frac{1}{n} \sum_{i=1}^n W_{t+1 \Rightarrow t}^{(i)}\left\|D_{t+1}^{(i)}-\hat{D}_t^{(i)}\right\|_1 \\
O P W=\sum_{t=0}^{T-1} O P W_t,
\end{gathered}
\end{equation}
where, $W_{t+1 \Rightarrow t}^{(i)}$ is optical flow based visibility mask calculated from the warping discrepancy between subsequent frames as explained in \cite{wang2022less}.

\clearpage
\section{Additional Qualitative Results}
\subsection{More Qualitative Temporal Consistency Comparisons} 
\begin{figure}[t!]
	\centering
	\includegraphics[width=0.9\linewidth]{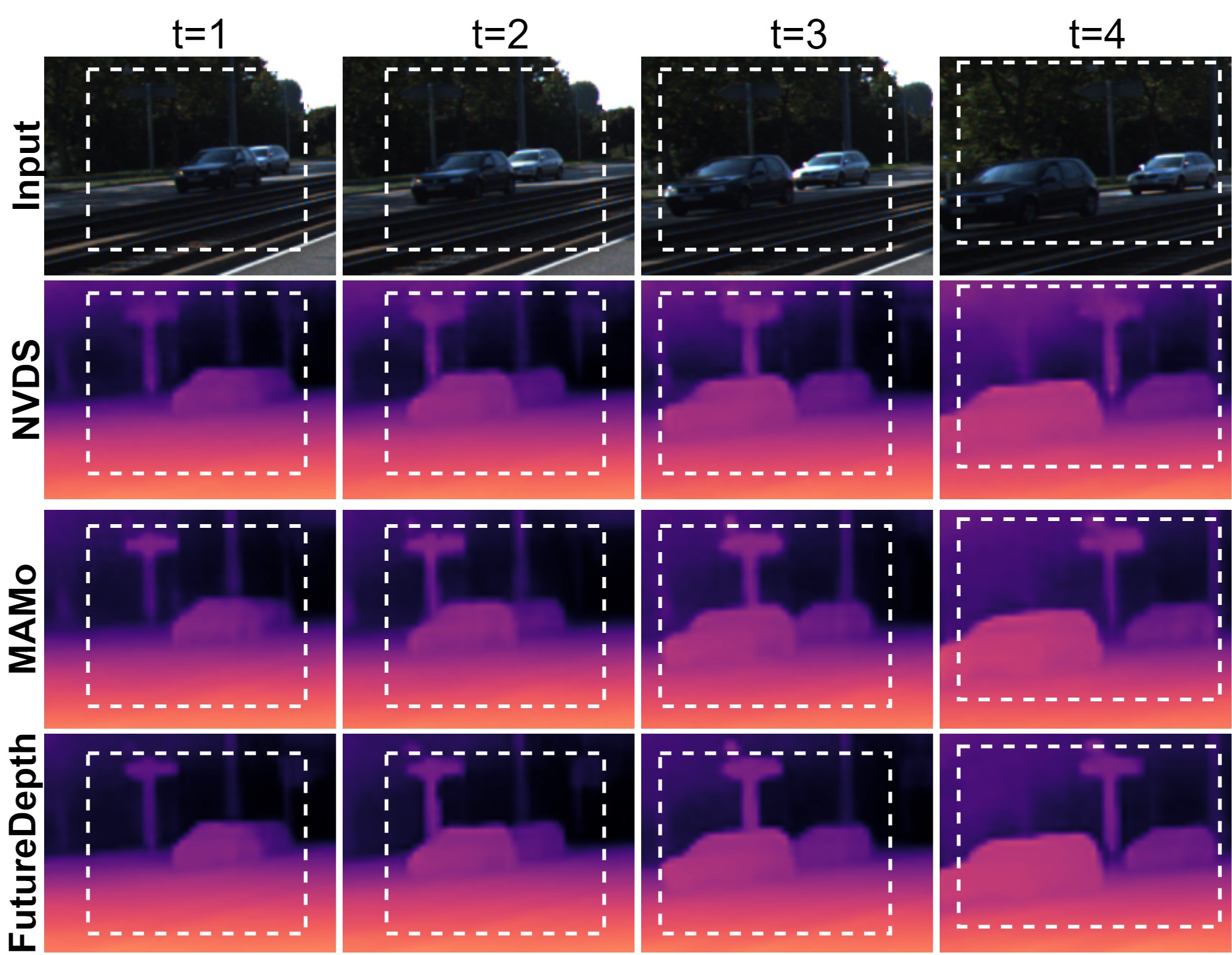} 
	\vskip -5pt 
	\caption{\footnotesize Sample patches from 4 consecutive frames. \ours~is more temporally consistent and accurate than existing SOTA (e.g., see that depths over the traffic sign are less accurately predicted in the cases of NVDS and MAMo, in the 4th frame after it is unoccluded). }
	\label{fig:time_seq-2}
	\vspace{-5pt}
\end{figure}

\begin{figure}[t!]
	\centering
	\includegraphics[width=0.9\linewidth]{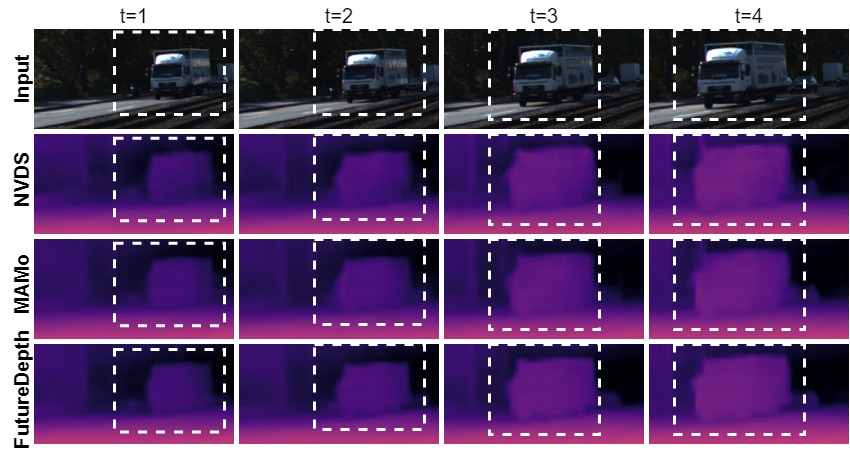} 
	\vskip -5pt 
	\caption{\footnotesize Sample patches from 4 consecutive frames. \ours~is more temporally consistent and accurate than existing SOTA (e.g., see that the front of the truck is more blurry in the cases of NVDS and MAMo). }
	\label{fig:time_seq-3}
	\vspace{-5pt}
\end{figure}

 Fig.~\ref{fig:time_seq-2} and~\ref{fig:time_seq-3} show visual comparisons of predicted depths on consecutive frames by \ours~and SOTA video depth estimation methods. We see that the depth predictions by MAMo and NVDS are inconsistent and noisy across frames, whereas our prediction is more temporally consistent and accurate.
 
\subsection{More Visualization Results of $Q_{scene}$} 
\begin{figure*}[h!]
	\centering
	\includegraphics[width=\linewidth]{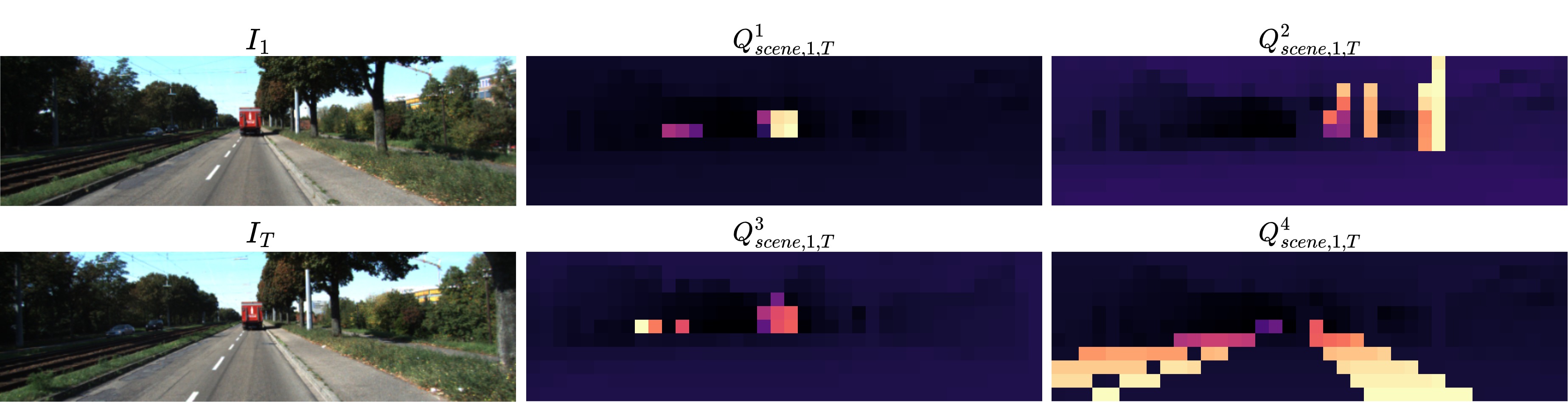} 
	\vskip -8pt 
	\caption{\small Sample $Q_{scene}$ generated by R-net. We show four sample channels in $Q_{scene,1,T}$ for input frames $I_{1,T}\, (T=4)$. We can clearly observe that important fore-ground and back-ground objects are captured as queries in $Q_{scene,1,T}$ that helps \ours~decoder in computing high quality depth maps. }
	\label{fig:example-qscene-1}
    \vspace{-10pt}
\end{figure*}

Fig.~\ref{fig:example-qscene-1} shows samples $Q_{scene}$ generated by R-net. We can clearly observe that important fore-ground and back-ground objects are captured as queries in $Q_{scene,1,T}$ that helps \ours~decoder in computing high quality depth maps.

\subsection{More Qualitative Results on Depth Estimation Quality}
\begin{figure*}[h!]
	\centering
	\includegraphics[height=0.9\textheight]{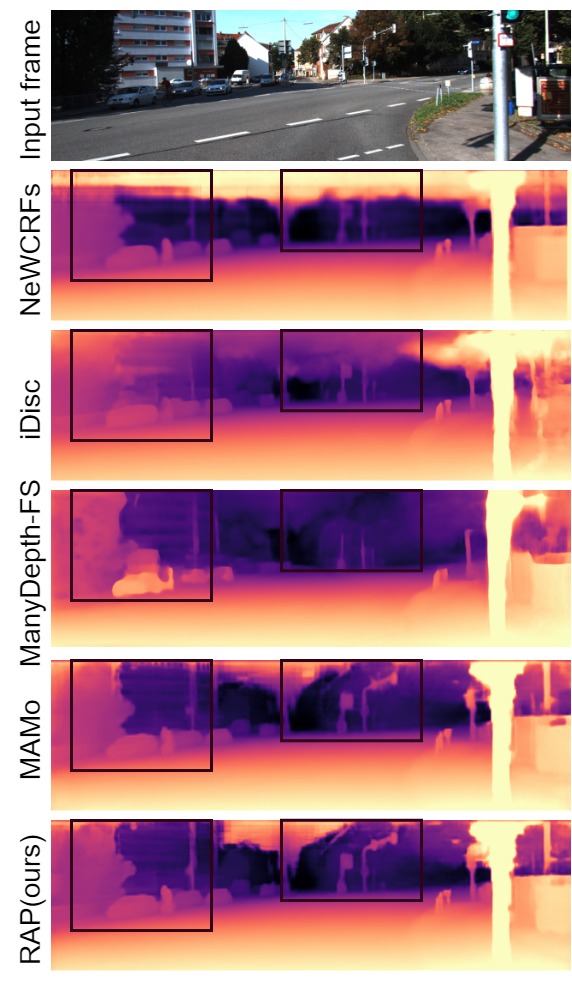} 
	\vskip -8pt 
	\caption{\small Qualitative results on KITTI. }
	\label{fig:visualizations-7}
	\vspace{-0pt}
\end{figure*}

\begin{figure*}[h!]
	\centering
	\includegraphics[height=0.9\textheight]{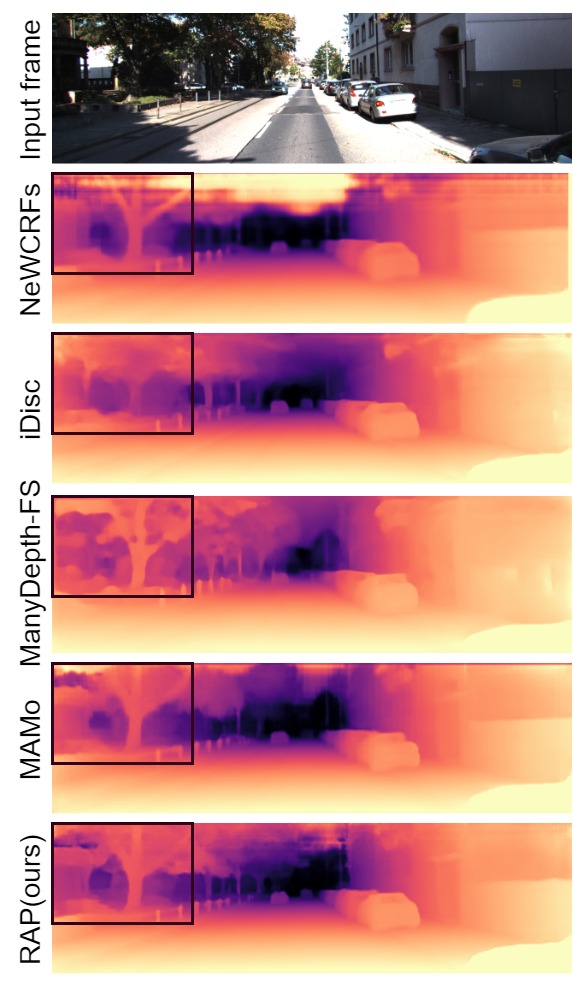} 
	\vskip -8pt 
	\caption{\small Qualitative results on KITTI. }
	\label{fig:visualizations-8}
	\vspace{-0pt}
\end{figure*}

\begin{figure*}[h!]
	\centering
	\includegraphics[height=0.9\textheight]{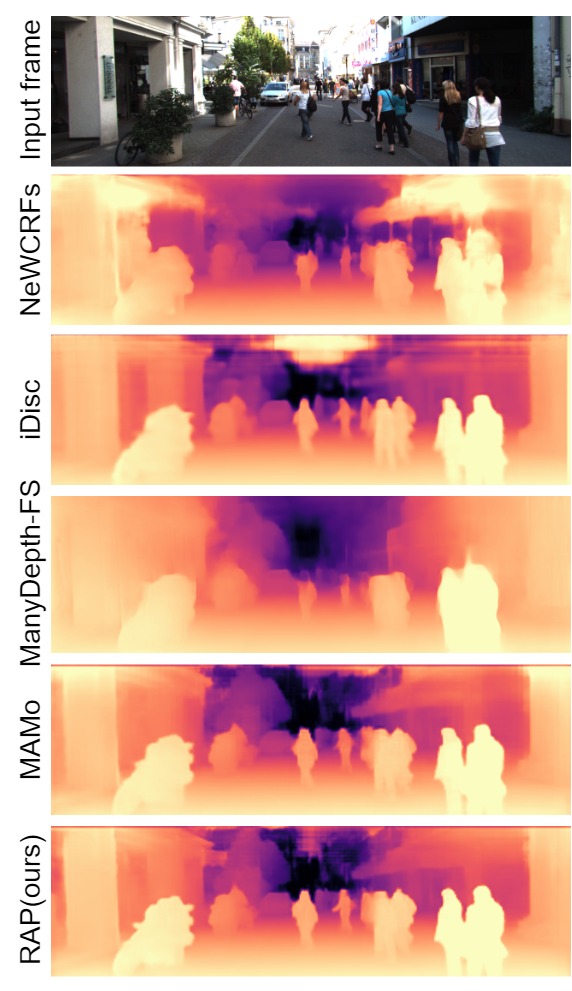} 
	\vskip -8pt 
	\caption{\small Qualitative results on KITTI. }
	\label{fig:visualizations-9}
	\vspace{-0pt}
\end{figure*}

Fig.~\ref{fig:visualizations-7}, ~\ref{fig:visualizations-8}, and ~\ref{fig:visualizations-9} provide more visual comparisons on depth quality between \ours~and existing SOTA methods. We see that our depth predictions are more accurate and better capture fine details.



\end{document}